%% file: main.tex
\documentclass{article}

\usepackage{microtype}
\usepackage[pdftex]{graphicx}
\usepackage{booktabs} 

\usepackage{hyperref}

\usepackage[accepted]{icml2024}

\usepackage{amsmath}
\usepackage{amssymb}
\usepackage{mathtools}
\usepackage{amsthm}

\usepackage{amsmath,amsfonts}
\usepackage{algorithmic}
\usepackage{graphicx}
\usepackage{textcomp}
\usepackage{array,colortbl,xcolor}

\usepackage{tkz-graph}

\usepackage{pgfplots}

\usepackage{tikz}
\usetikzlibrary{decorations.pathreplacing,calligraphy}
\usetikzlibrary{patterns}
\usetikzlibrary{arrows}
\usetikzlibrary{calc}
\usetikzlibrary{shapes}

\usepackage{pgfkeys}
\usetikzlibrary{backgrounds}
\pgfplotsset{compat=newest}
\usepackage[export]{adjustbox}
    
\usepackage{comment}
    
\usepackage{xspace}

\usepackage{algorithm}
\usepackage{algorithmic,multicol}
\usepackage{multirow}

\usepackage{booktabs}

\usepackage{tabularx}

\usepackage{pdfpages}

\usepackage{listings}

\usepackage{dsfont}

\usepackage{caption}
\usepackage{subcaption}

\usepackage[capitalize,noabbrev]{cleveref}

\theoremstyle{plain}

\theoremstyle{definition}

\theoremstyle{remark}

\input{tex/macros}

\usepackage[textsize=tiny]{todonotes}

\icmltitlerunning{\qute: Quantifying Uncertainty in TinyML models with Early-exit assisted ensemble}

\begin{document}

\twocolumn[
\icmltitle{\qute: Quantifying Uncertainty in TinyML models with Early-exit-assisted ensembles for model-monitoring}



\icmlsetsymbol{equal}{*}

\begin{icmlauthorlist}
\icmlauthor{Nikhil P Ghanathe}{yyy}
\icmlauthor{Steven J E Wilton}{yyy}
\end{icmlauthorlist}

\icmlaffiliation{yyy}{Department of Electrical and Computer Engineering, University of British Columbia, Vancouver, Canada}

\icmlcorrespondingauthor{Nikhil P Ghanathe}{nikhilghanathe@ece.ubc.ca}

\icmlkeywords{Machine Learning, ICML}

\vskip 0.3in
]



\printAffiliationsAndNotice{\icmlEqualContribution} 

\input{tex/0-abstract}

\input{tex/1-intro}

\input{tex/2-related}
\input{tex/3-background}
\input{tex/4-design}
\input{tex/5-methodology}

\input{tex/6-results}

\input{tex/7-conclusion}


\nocite{langley00}

\bibliography{bibliography}
\bibliographystyle{icml2024}

\newpage
\appendix
\onecolumn
\input{tex/9-appendix}

\end{document}

%% file: tex/macros.tex
\newcommand{\qute}{\textsc{QUTE}\xspace}
\newcommand{\bigmcu}{\textsf{Big-MCU}\xspace}
\newcommand{\smallmcu}{\textsf{Small-MCU}\xspace}

\definecolor{bblue}{HTML}{4F81BD}
\definecolor{rred}{HTML}{C0504D}
\definecolor{ggreen}{HTML}{9BBB59}
\definecolor{ppurple}{HTML}{9F4C7C}
\definecolor{yyellow}{HTML}{DDAE06}

\tikzstyle{bar1} = [bblue, fill=bblue, postaction={pattern=north east lines}]
\tikzstyle{bar2} = [rred, fill=rred]
\tikzstyle{bar3} = [ggreen, fill=ggreen, postaction={pattern=north west lines}]
\tikzstyle{bar4} = [ppurple, fill=ppurple, postaction={pattern=crosshatch dots}]
\tikzstyle{bar5} = [yyellow, fill=yyellow, postaction={pattern=grid}]

\newcommand{\graphHeight}{0.15\textheight}

%% file: tex/0-abstract.tex
\begin{abstract}
Uncertainty quantification (UQ) provides a resource-efficient solution for on-device monitoring of tinyML models deployed without access to true labels. However, existing UQ methods impose significant memory and compute demands, making them impractical for ultra-low-power, KB-sized TinyML devices. Prior work has attempted to reduce overhead by using early-exit ensembles to quantify uncertainty in a single forward pass, but these approaches still carry prohibitive costs. To address this, we propose \qute, a novel resource-efficient early-exit-assisted ensemble architecture optimized for tinyML models. \qute introduces additional output blocks at the final exit of the base network, distilling early-exit knowledge into these blocks to form a diverse yet lightweight ensemble. We show that \qute delivers superior uncertainty quality on tiny models, achieving comparable performance on larger models with \textbf{59\% smaller model sizes} than the closest prior work. When deployed on a microcontroller, \qute demonstrates a \textbf{31\% reduction in latency} on average. In addition, we show that \qute excels at detecting accuracy-drop events, outperforming all prior works. 

\end{abstract}

%% file: tex/1-intro.tex
\section{Introduction}
\label{sec:intro}

Recent advancements in embedded systems and machine learning (ML) have produced a new class of edge devices containing powerful ML models. These milliwatt-scale KB-sized devices, often termed \emph{TinyML} devices, have low compute and memory requirements. They are often deployed in  remote environments with no availability of true labels. This makes them susceptible to both out-of-distribution (OOD) and covariate-shifted data caused by environmental and sensor variations that manifest often unpredictably in the field. 
The ability of tinyML models to accurately measure the uncertainty of their predictions is crucial for two reasons. First, these devices generate data that is frequently used in critical decision-making.  In the context of an autonomous vehicle (AV), uncertain predictions could result in the downstream system making cautious driving decisions~\cite{tang2022-uncertainty-in-av}. Second, tinyML devices often operate in remote environments~\cite{vargas2021-overview-av-vulnerability}. If the inputs to the model changes (perhaps due to spatter, fog, frost, noise, motion blur, etc.), being aware of the model's unreliability may prompt an engineer to take remedial action. Many prior methods for drift/corruption detection are either statistical tests/methods that operate directly on input data~\cite{bifet2007learning-adwin} with huge memory requirements, or require true labels for error-rate calculation~\cite{gama2004-DDM, baier2022-detecting-drift-uncertainty}. In contrast, uncertainty quantification (UQ) methods aim to estimate a model's \textit{prediction confidence} without requiring true labels, providing an efficient solution for model monitoring, particularly in resource and power-constrained environments.

\begin{figure}[t!]
    \centering
    \vspace{-1.5cm}
    \scalebox{0.55}{\input{figs/ev-ensemble-horz}}
    \caption{\qute architecture. $\{f_{out}\}_{k=1}^{K}$ represents the 'K' additional output blocks at the final exit, which are \emph{assisted} by 'K' early-exit blocks $\{g_{\theta_k}\}_{k=1}^{K}$ \textit{only during training} to promote diversity (see Figure~\ref{fig:EV-assist}). For inference, all \emph{(early-exits \& $f_{out}$) are removed}\vspace{-0.5cm}}
    \label{fig:qute}
\end{figure}

In ML models, there are two main types of uncertainty~\cite{kendall2017-unceratainties-bayesian-DL}: \textit{epistemic} and \textit{aleatoric}. Epistemic uncertainty is \textit{reducible} and stems from limited data or knowledge, while aleatoric uncertainty is \textit{irreducible}, arising from inherent randomness in data or model limitations. In field-deployed tinyML models, both types of uncertainty can emerge due to semantic (OOD) and non-semantic changes in inputs. The non-semantic category, which we term \emph{corrupted-in-distribution} (CID) data, occurs when sensor malfunctions or environmental factors result in corrupted versions of expected data (e.g., fogged camera lens). CID data introduces both aleatoric (e.g., due to random noise or weather) and epistemic (e.g., from blur or digital corruptions) uncertainties, often simultaneously, making it crucial to capture these corruption-induced uncertainties to ensure model \textit{reliability}.
Unfortunately, modern neural networks are poor in estimating uncertainty of their predictions~\cite{ovadia2019-trust}. Many prior works have proposed uncertainty-aware networks: ensemble networks~\cite{lakshminarayanan2017-uncertainty-est-ensembles} have been found to be extremely effective in capturing both aleatoric and epistemic uncertainties.  
Alternatively, early-exit networks have been converted into ensembles~\cite{antoran-2020-DUN, qendro2021-EEensemble, ferianc-2023-MIMMO}. However, both these approaches incur high  memory/compute overheads and are untenable for tinyML. 

In this paper, we propose \qute (Figure~\ref{fig:qute}), a novel and resource-efficient early-exit-assisted ensemble architecture that enables high-quality uncertainty quantification in tinyML models in the context of both in-distribution (ID) and corrupted-in-distribution (CID) data.  
As shown in Figure~\ref{fig:qute}, we append additional lightweight classification heads to an existing base neural network to create ensemble members, and crucially, 
these classification heads are assisted by the early-exits in a manner similar to the Early-View assistance technique~\cite{ghanathe2023trecx} to promote diversity. Post-training, we eliminate the resource-hungry early-exits while retaining only the economical additional classification heads.
Our approach has significantly less memory and compute overhead  compared to prior works (59\% smaller models and 3.2$\times$ fewer FLOPS compared to the most relevant prior work~\cite{qendro2021-EEensemble}). Furthermore, \qute performs better than prior methods in estimating uncertainty caused due to CID error sources, and on-par with prior methods for uncertainty due to OOD.
We further show that higher uncertainty is correlated with a drop in accuracy. We evaluate \qute's ability to detect such accuracy drop events caused by CID against prior methods, and show that \qute outperforms all prior methods. 
To the best of our knowledge, this is the first early-exit ensemble architecture for uncertainty quantification optimized for tinyML models. 

This paper is organized as follows.
Section~\ref{sec:related} presents related work.  The context and problem formulation are  in 
Section~\ref{sec:background}.  Our approach is described in 
Section~\ref{sec:design}.  The experimental methodology and evaluation results
are in Sections~\ref{sec:methodology} and~\ref{sec:results} respectively.
Section~\ref{sec:conclusion} concludes the paper.

%% file: figs/ev-ensemble-horz.tex
\begin{tikzpicture}[font=\sffamily\small\fontfamily{LinuxBiolinumT-LF}\selectfont\Large,>=stealth']

  \tikzset{ionode/.style={font=\sffamily\small\fontfamily{LinuxBiolinumT-LF}\selectfont\large,>=stealth', align=center,minimum width=1.5cm,black!75}};
  
  \tikzset{ionodebig/.style={font=\sffamily\small\fontfamily{LinuxBiolinumT-LF}\selectfont\Large,>=stealth', align=center,minimum width=1.5cm,black!75}};
  
  \tikzset{layernode/.style={align=center,draw,minimum height=1cm, minimum width=1.5cm, rounded corners=3pt,inner sep=5pt,fill=black!5}};
  
  \tikzset{layernode_dotted/.style={dotted, align=center,draw,minimum height=1cm, minimum width=1.5cm, rounded corners=3pt,inner sep=5pt,fill=black!5}};
  
  \tikzset{classfiernode/.style={align=center,draw,minimum height=3cm,rounded corners=4mm,inner xsep=8pt,, fill=yellow}};
  
  \tikzset{poolingnode/.style={align=center,draw,minimum height=2.5cm,rounded corners=5mm,inner xsep=8pt,fill=blue}};
  
  \tikzset{circlenode/.style={align=center,draw,minimum height=0.5cm,rounded corners=5mm,inner xsep=4pt,fill=blue}};
  
  \tikzset{eenode/.style={dotted, thick, align=center,draw, minimum height=0.7cm, minimum width=2cm, rounded corners=1mm,inner xsep=8pt,fill=red}};

  \node (anchor) [ionode,black] {};
  \node (input_block) at (anchor.east) [ionode, black, xshift=-5cm, yshift=0.0cm] {\textbf{Input ($f_0$)}};
  \node (f1) at (input_block.south) [layernode, black, xshift=0cm, yshift=-1.0cm, fill=black!10] {$f_1$};
  \node (f2) at (f1.east) [layernode, black, xshift=1.5cm, yshift=0.0cm, fill=black!10] {$f_2$};
  \node (fd_1) at (f2.east) [layernode, black, xshift=2.5cm, yshift=0.0cm, fill=black!10] {$f_{D-1}$};
  \node (fd) at (fd_1.east) [layernode, black, xshift=1.5cm, yshift=0.0cm, fill=black!10] {$f_{D}$};

    \node (fout0) at (fd.east) [layernode_dotted, black, xshift=1.5cm, yshift=1.5cm, fill=yellow!40] {$f_{out}$};
  \node (foutk) at (fd.east) [layernode, black, xshift=1.5cm, yshift=0.2cm, fill=yellow!15] {$f_{out_k}$};
  \node (fout2) at (fd.east) [layernode, black, xshift=1.5cm, yshift=-2.1cm, fill=yellow!15] {$f_{out_2}$};
  \node (fout1) at (fd.east) [layernode, black, xshift=1.5cm, yshift=-3.5cm, fill=yellow!15] {$f_{out_1}$};


  \node (g1) at (f1.south) [eenode,anchor=west,xshift=0.0cm, yshift=-1cm, fill=red!15] {$g_{{\theta}_1}$};
  \node (g2) at (f2.south) [eenode,anchor=west,xshift=0.0cm, yshift=-1cm, fill=red!15] {$g_{{\theta}_2}$};
  \node (gd_1) at (fd_1.south) [eenode,anchor=west,xshift=0.0cm, yshift=-1cm, fill=red!15] {$g_{{\theta}_{D-1}}$};

  \node (pd_fout0) at (fout0.east) [ionodebig, black, xshift=1.2cm, yshift=-00cm] {$p_{\phi}(y|x)$};
  \node (pd_fout1) at (fout1.east) [ionodebig, black, xshift=1.2cm, yshift=0.0cm] {$p_{{\phi}_1}(y|x)$};
  \node (pd_fout2) at (fout2.east) [ionodebig, black, xshift=1.2cm, yshift=0.0cm] {$p_{{\phi}_2}(y|x)$};
  \node (pd_foutk) at (foutk.east) [ionodebig, black, xshift=1.2cm, yshift=0.0cm] {$p_{{\phi}_k}(y|x)$};
  
  \node (ee1) at (g1.south) [ionodebig, black, xshift=0cm, yshift=-1cm] {$p_{{\theta}_1}(y|x)$};
  \node (ee2) at (g2.south) [ionodebig, black, xshift=0cm, yshift=-1cm] {$p_{{\theta}_2}(y|x)$};
  \node (eed_1) at (gd_1.south) [ionodebig, black, xshift=0cm, yshift=-1cm] {$p_{{\theta}_{D-1}}(y|x)$};
  
  \node (depth) at (f2.east) [ionodebig, black, xshift=1.0cm, yshift=0.4cm] {$\textbf{D}$};

  \draw[->] (input_block) -- (f1);
  \draw[->] (f1) -- (f2);
  \draw[->, thick] (f2.east) -- ++(0.5cm,0cm);
  \draw[->, dotted, thick] (f2.east)  -- (fd_1);
  
  \draw[->] (fd_1) -- (fd);
  \draw[->] (fd) -- (fd.east) -- ++(0.3, -0.0cm) -- ++(0cm, 1.5cm) -- ++(0.4cm, 0.0cm);
  \draw[->] (fd) --  (fd.east) -- ++(0.3, -0.0cm) -- ++(0cm, 0.2cm) -- ++(0.4cm, 0.0cm);
  \draw[->] (fd) --  (fd.east) -- ++(0.3, -0.0cm) -- ++(0cm, -2.1cm) -- ++(0.4cm, 0.0cm);
  \draw[->] (fd) --  (fd.east) -- ++(0.3, -0.0cm) -- ++(0cm, -3.5cm) -- ++(0.4cm, 0.0cm);
  
  \draw[->] (fout0) -- (pd_fout0);
  \draw[->] (fout1) -- (pd_fout1);
  \draw[->] (fout2) -- (pd_fout2);
  \draw[->] (foutk) -- (pd_foutk);

  \draw[->] (f1) -- (f1.south) -- ++(0, -0.2cm) -- ++(1cm, 0) -- ++(0, -0.4cm);
  \draw[->] (f2) -- (f2.south) -- ++(0, -0.2cm) -- ++(1cm, 0) -- ++(0, -0.4cm);
  \draw[->] (fd_1) -- (fd_1.south) -- ++(0, -0.2cm) -- ++(1.2cm, 0) -- ++(0, -0.4cm);

  \draw[->] (g1) -- (ee1);
  \draw[->] (g2) -- (ee2);
  \draw[->] (gd_1) -- (eed_1);

  \draw[dotted, ultra thick] (fout2) ++(0cm, 0.5cm) -- ++(0, 1.3cm);
  \draw[dotted, ultra thick] (ee2) ++(1cm, 0cm) -- ++(1cm, 0cm);
  \draw[dotted, ultra thick] (g2) ++(1.2cm, 0cm) -- ++(0.9cm, 0cm);



\draw [decorate,decoration={brace,amplitude=10pt,raise=4pt},yshift=0pt]
(foutk.east)+(2.0cm, 0.3cm) -- ++(2.0cm, -4cm) node [black,midway, text width=1cm, xshift=1.2cm, yshift=0cm]{$\mathcal{K}$};

\draw [decorate,decoration={mirror, brace,amplitude=10pt,raise=4pt},yshift=0pt]
(ee1.west)+(-0.2cm, -0.5cm) -- ++(7.7cm, -0.5cm) node [black,midway, text width=10cm, xshift=0cm, yshift=-1.3cm]{Resource-hungry EEs \textit{assist} the output blocks $\{f_{out_k}\}_{k=1}^K$ only during training, promoting diversity. \textit{Removed for inference}};

\draw []
(foutk.west)+(-1.5cm, 4.5cm) node [black,midway, text width=11cm, xshift=1cm, yshift=-7.5cm] (ensemble) {\textsc{Ensemble $(\mathcal{K}) = \{p_{{\phi}_1}(y|x), p_{{\phi}_2}(y|x),...., p_{{\phi}_K}(y|x)\}$}}; 
\draw [black] (ensemble.south west) rectangle (ensemble.north east);



  
\end{tikzpicture}

%% file: tex/2-related.tex
\section{Related Work}
\label{sec:related}
\textbf{Uncertainty quantification}~Bayesian neural networks (BNN) are well-suited to quantify uncertainty of a model~\cite{blundell2015-weight-uncertainty, hernándezlobato2015-bayesian-1, teye2018-bayesian-2}, but are parameter-inefficient and incur a high resource/compute overhead. Monte Carlo Dropout (MCD)~\cite{gal2016-MCdropout} creates implicit ensembles by enabling dropout during multiple inference passes.
Recently, ensemble networks like DeepEnsembles~\cite{lakshminarayanan2017-uncertainty-est-ensembles} have been shown to produce good uncertainty estimates~\cite{zaidi2020-neural-ensembles, wenzel2020-hyper-deep-ensembles, rahaman2021-uncertainty-quantification-ensembles}. 
However, this too requires multiple inferences of individual networks, and is impractical in terms of memory for tinyML. 
Other prior works have proposed multi-input and multi-output networks that combine multiple independent networks into one~\cite{havasi2021-MIMO, ferianc-2023-MIMMO}, but they do not scale well and remain impractical for tinyML.~\cite{ahmed2024-tinydeepensemble} ensembles only normalization layers, but requires specialized hardware.
Alternatively, prior works~\cite{qendro2021-EEensemble, ferianc-2023-MIMMO} have leveraged early-exit networks to create implicit ensembles. The closest work to \qute is EE-ensemble~\cite{qendro2021-EEensemble}, which uses outputs of early-exits as ensemble members. However, the early-exits appended with extra learning layers incur a prohibitive cost (Section~\ref{sec:background}).  In contrast, ensemble distillation methods~\cite{havasi2021-MIMO, malinin2019-ensemble-dist-distill, tran2020-hydra} filter the knowledge of all ensemble members into a conventional neural network (NN). Hydra~\cite{tran2020-hydra}, which also uses a single multi-headed network is closest to our work.


\textbf{Single-pass deterministic network}~Several single-pass (non-Bayesian) methods for UQ~\cite{van2020-duq, mukhoti2023-ddu, sensoy2018-edl, deng2023-fisher} offer lower memory footprints compared to ensemble methods. However, they remain impractical for tinyML due to specialized output layers or architectural constraints. DUQ~\cite{van2020-duq} adds a specialized layer post-softmax, drastically increasing resource use (e.g., 10$\times$ more parameters for ResNet on CIFAR-10). DDU~\cite{mukhoti2023-ddu} simplifies this but relies on residual connections for feature space regularization, limiting its applicability. Priornets~\cite{malinin2018-priornets} require OOD data, which is often unrealistic. Postnets~\cite{charpentier2020-postnets}, though OOD-free, focus primarily on OOD detection, often sacrificing accuracy below the base network's level. Other methods like~\cite{meronen2023-fixingoverconfidencedynamicneural} that address overconfidence in early-exit networks require excessive computational resources (e.g., laplace approximations).


\textbf{Model monitoring}~There has been a plethora of work to detect OOD samples, which represents a semantic shift from in-distribution ~\cite{yang2021-generalizedOODsurvey, yang2022-openood, zhang2023-survey-learning-to-reject, liu2021-energybasedood, hendrycks2018-baseline-for-ood, dinari2022-vae-ood} . However, there have been fewer works that deal with detecting corrupted-ID, which represent a non-semantic shift w.r.t. ID, and are much harder to detect~\cite{liang2020-odin}. Some prior works utilize corrupted-ID/covariate-shifted ID to better generalize on OOD~\cite{bai2023-feed-two-birds, katzsamuels2022-training-ood-in-natural-habitats}, which is difficult to obtain.
The most relevant work  that leverages \emph{only} training data is generalized-ODIN (G-ODIN)~\cite{hsu2020-generalized-odin}. It adds 1)~a preprocessing layer that perturbs the input image and 2)~decomposes confidence score for better OOD detection. 
However, the preprocessing might prove costly/impractical on tinyML devices. Recent studies~\cite{xia2023-failure-detection, jaeger2023-failure-detection} propose \textit{failure detection} with a \textit{reject} option, rejecting high-uncertainty samples from both ID (potentially incorrect predictions) and OOD, unlike traditional OOD detection which only separates ID from OOD.

\textbf{Early-exit networks}~Early-exit networks add intermediate exits along the length of the base network thereby, providing avenues for reduction in average inference time~\cite{branchynet, kaya2019-sdn, msdnet, class-spec, ghanathe2023trecx, jazbec2024-anytime-ee}. 
~\citet{ghanathe2023trecx} introduces an early-exit architecture optimized for tinyML models. In addition, it introduces the \emph{early-view assistance} method, primarily to mitigate network overthinking in neural networks to boost accuracy. In contrast, we utilize the underlying principle and develop a modified version of this method in our work to principally create a diverse ensemble (Section~\ref{sec:design}).  

%% file: tex/3-background.tex
\section{Background and Problem formulation}
\label{sec:background}

    

    
\input{figs/table-resnetstacks}

Consider an \emph{in-distribution} dataset $\mathrm{S_{ID}}=\{x_n, y_n\}_{n=1}^{N}$ of size $N$  where, $x_n$ and $y_n$ are the n\textsuperscript{th} input sample and its corresponding true label respectively. A discriminative model $\mathcal{M}_\Theta(x)$ learns parameters $\Theta$ on $\mathrm{S_{ID}}$ and outputs a class posterior probability vector $p_\Theta(y|x)$, which yields a predicted label $\hat{y}$. For a classification problem, $\hat{y}\in\{1,2,..L\}$, where $L$ is the number of classes. Uncertainty quantification methods endeavor to improve uncertainty estimation quality of $\mathcal{M}_\Theta(x)$ on $\mathrm{S}_{ID}$.  
Specifically, we want to calibrate the model such that its \emph{predictive confidence}
($\mathcal{C}_{\mathcal{M}_\Theta}$) is in sync with its accuracy ($\mathcal{A}_{\mathcal{M}_\Theta}$). 
A well-calibrated model will see a commensurate drop in  $\mathcal{C}_{\mathcal{M}_\Theta}$  as $\mathcal{A}_{\mathcal{M}_\Theta}$ drops. The predictive confidence is given by, $\mathcal{C}_{\mathcal{M}_\Theta}(x) = \underset{{l\in \{1,2,..,L}\}}{max}  ~~p_\Theta(y=l|x)$
Confidence-calibration is a well studied problem, and it helps achieve synergy between the predicted probabilities and ground truth correctness likelihood~\cite{guo2017-ece}.
However, when $\mathcal{M}_\Theta$ is deployed in a real-world scenario, it may encounter either 1)~out-of-distribution data ($\mathrm{S_{OOD}})$ or 2)~a corrupted/covariate-shifted version ($\mathrm{S_{CID}}$), due to environmental/sensor variations (E.g., frost, noise, fog, rain etc.). 
Unfortunately, we find that despite good uncertainty estimation quality on $\mathrm{S_{ID}}$, many uncertainty-aware networks see a drop in quality and remain overconfident, particularly for severely-corrupted $\mathrm{S}_{CID}$~\cite{hsu2020-generalized-odin}. Interestingly, we find that the model becomes \emph{less overconfident} to $\mathrm{S_{CID}}$ as its \emph{size shrinks}. 
This phenomenon is illustrated in Table~\ref{table:uq-stack}. We report the calibration metrics of Resnet~\cite{resnetv1} on CIFAR10, with four model sizes ranging from 1 to 4 residual stacks on both $\mathrm{S_{ID}}$ and $\mathrm{S_{CID}}$. For calibration metrics (ECE, BS, NLL), \textit{lower is better}. More details on calibration metrics and CIFAR10-CID can be found in Section~\ref{sec:methodology} and  Appendix~\ref{sec:apdx-metrics}.
As seen, the 4stack model with the highest model capacity is clearly the best on $\mathrm{S_{ID}}$. However, on $\mathrm{S_{CID}}$, it performs the worst, remaining overconfident in the presence of corruptions, even outperformed by 1stack. In contrast, 2stack and 3stack model are much better on $\mathrm{S_{CID}}$. This demonstrates that smaller models are \emph{less overconfident} for corruption in inputs, leading to better calibration on $\mathrm{S_{CID}}$. 


The capabilities of smaller models can be harnessed for better uncertainty estimation on $\mathrm{S_{CID}}$ through early-exit networks, as illustrated by prior research like EE-ensemble~\cite{qendro2021-EEensemble}, which combines the predictions from multiple early-exits and final exit to form an ensemble. For example, in our evaluations, we find that on TinyImagenet-ID~\cite{le2015-tinyimagenet} with MobilenetV2~\cite{mobilenets}, the negative log-likelihood (NLL) of EE-ensemble is \emph{26\% lower} than the popular DeepEnsemble~\cite{lakshminarayanan2017-uncertainty-est-ensembles}, and \emph{34\% lower} on TinyImagenet-Corrupted~\cite{hendrycks2019-benchmarking-corruptions} (lower is better). This showcases the crucial role of early-exits in achieving higher quality uncertainty overall. 

However, we find that EE-ensemble is resource-intensive because it adds additional learning layers at the early-exits to accommodate their varying learning capacities. For example, adding two early-exits \emph{without additional learning layers} (say EE-0 and EE-1) after 1\textsuperscript{st} and 2\textsuperscript{nd} residual stack of a 3-residual stack Resnet for CIFAR10 results in individual accuracies of 0.602 and 0.707 at EE-0 and EE-1 respectively. This is significantly worse than the individual accuracy of the final exit, 0.84, which in turn leads to poor ensemble behavior. To address this, EE-ensemble uses early-exits \emph{with additional dense/fully-connected layers}, which incurs a high memory overhead  affecting practicality in tinyML (Overhead\textsubscript{EE-ensemble} = 397K params compared to Overhead\textsubscript{\qute} = 8.2K params as shown in Section~\ref{sec:design} 
\&~\ref{sec:results}).
We present an alternative strategy to leverage the knowledge of early-exits in ensemble-creation that is extremely resource-efficient thereby enabling superior uncertainty estimates on both $\mathrm{S}_{ID}$ and $\mathrm{S}_{CID}$, which allows us to reliably detect accuracy-drops in the model due to $\mathrm{S}_{CID}$.

%% file: figs/table-resnetstacks.tex

\begin{table}[t]
\centering
\small
\scalebox{0.65}{

{
\begin{tabular}{l | c c c c || c c c c}
  &  \multicolumn{4}{c||} {\textit{In-distribution data}} & \multicolumn{4}{c}{\textit{Corrupted-in-distribution data}}\\
\hline
\textbf{Model}  & \textbf{Acc ($\uparrow$)}  & \textbf{ECE ($\downarrow$)} & \textbf{BS ($\downarrow$)} & \textbf{NLL ($\downarrow$)} &     \textbf{Acc ($\uparrow$)}  & \textbf{ECE ($\downarrow$)} & \textbf{BS ($\downarrow$)} & \textbf{NLL ($\downarrow$)} \\
\hline

 ~-~1stack &     0.72 & 	0.0343 & 	0.0381 & 	0.8174 & 	 	0.204 & 	0.2796 & 	0.0993 & 	3.216\\
 
~-~2stack &     0.78 & 	\textbf{0.0315} & 	0.0305 & 	0.6441 & 	 	0.207 & 	\textbf{0.2518} & 	0.0969 & 	\textbf{3.158}\\

~-~3stack &    0.87 & 	0.0373 & 	0.01907 & 	0.4063 & 	 	\textbf{0.27} & 	0.2534 & 	\textbf{0.0953} & 	3.201\\

~-~4stack &    \textbf{0.89} & 	0.0562 & 	\textbf{0.01615} & 	\textbf{0.4019} & 	 	0.25 & 	0.4255 & 	0.1087 & 	3.677\\

\end{tabular}
}
}
\caption{Comparing calibration Metrics of Resnet with \{1,2,3,4\} stack models on CIFAR10 ID and CID. \emph{Best results marked in bold}. For ECE, BS, NLL \emph{lower is better}\vspace{-0.5cm}}
\label{table:uq-stack}
\end{table}

%% file: tex/4-design.tex
\section{\qute}
\label{sec:design}
A neural network (NN) like the one shown in Figure~\ref{fig:qute}, with depth $D$ is composed of several blocks of linear/non-linear operations (e.g., convolution, pooling) that are stacked. The NN consists of $D-1$ intermediate blocks $\{f_i(.)\}_{i=1}^{D-1}; f(.) = f_1(.) \circ f_2(.) \circ f_3(.) .... \circ f_{D-1}(.) \circ f_{D}(.)$ and an output block $f_{out}(.)$. $f_0(.)$ is the input block. 
In the base network, the network processes the input $f_0(.)$ through each block $f_i(.)$ until layer $D$, ultimately producing prediction $p_{{\Theta}}(y|x)$ through output block $f_{out}(.)$.

Given a base NN (the grey blocks in Figure~\ref{fig:qute}), we first create $K$ additional classification heads/output blocks $\{f_{{out}_k}\}_{k=1}^{K}$ at the final exit (after $f_D$) as shown in Figure~\ref{fig:qute}. These output blocks constitute the ensemble $\mathcal{K}$. Next, we add $K$ early-exit blocks, $\{g_{{\theta}_k}\}_{k=1}^{K}$ along the length of the base network, which are used \textit{only during training} for knowledge distillation. A more intuitive view of \qute is presented in the Appendix.
In Figure~\ref{fig:qute}, $K$ is set to $D-1$. In practice, $K$ is a hyperparameter that depends on computation/resource budget and the required ensemble size $|\mathcal{K}|$, where the ensemble size is bounded above by the depth of the network. Thus, $\{f_{{out}_k}\}_{k=1}^{K}$ produces $K$ predictions, where each $f_{out_k}$ is assisted in producing predictions by its corresponding early-exit $g_{\theta_k}$ using \emph{early-view assistance} method.
\vspace{-1mm}
\begin{eqnarray}
    \mathcal{K} = \{p_{{\phi}_1}(y|x), p_{{\phi}_2}(y|x), p_{{\phi}_3}(y|x),...., p_{{\phi}_K}(y|x)\}\label{eq:ensemble}
\end{eqnarray}


Figure~\ref{fig:EV-assist} illustrates the architecture of \emph{early-view assistance}, where a single final exit block $f_{{out}_k}$ and its \emph{assisting} early-exit $g_{{\theta}_k}$ is shown. As shown, we create an additional learning layer $h_{{\phi}_k}(.)$ at the $k$\textsuperscript{th} output block. An identical learning layer $h_{{\theta}_k}(.)$ is also added at the $k$\textsuperscript{th} early-exit. $h_{{\phi}_k}$ with parameters $\phi_k$ is responsible for assimilating knowledge from the corresponding $h_{{\theta}_k}$ with parameters $\theta_k$. $\sigma(.)$ is the output activation (E.g., dense+softmax). $h_{{\phi}_k}$ and $h_{{\theta}_k}$ are a single depthwise-convolution layer in our evaluations owing to its low computation and memory demand. 
We ensure equal dimensionality between $h_{{\phi}_k}$ and $h_{{\theta}_k}$ by adding a pointwise-convolution layer before $h_{{\theta}_k}$ at all early-exits (not shown in figure for brevity). 
Fundamentally, our aim is to imbue the diverse predictive behaviors of early-exits from different depths into the output blocks $\{f_{out_k}\}_{k=1}^{K}$.

\begin{figure}[t]
    \centering
    \hspace{-0.7cm}
    \scalebox{0.6}{\input{figs/ev-assist-horz}}
    \centering\caption{Early-view-assistance architecture. One \emph{assisting} early-exit $g_{\theta_k}$ and corresponding \emph{early-view-assisted} exit $f_{out_k}$ is shown. Early-exits weights $\theta_k$ are transferred/copied to $h_{\phi_k}$ before each train batch, i.e., $\phi_k = \theta_k$}  
\label{fig:EV-assist}
\end{figure}

\textbf{Training}~
Unlike EE-ensemble, all output blocks of \qute have similar learning capacities since the input to all $\{h_{{\phi}_k}\}_{k=1}^{K}$ is $f_{D}$ of the base network. The output of $h_{{\phi}_k}$ is termed as \emph{early-view} since it is assisted by the early-exit.
After attaching early-exits (EE) and early-view-assisted (EV) exits to the untrained base network, all weights are learned simultaneously during training. Unlike some prior works~\cite{lakshminarayanan2017-uncertainty-est-ensembles} that use extra data (e.g., adversarial samples), we only use the available training data. We use random data augmentation (E.g., random rotation/flip/crop), a well-known method to improve model robustness to corruptions to ensure fairness in accuracy-drop detection experiments (Section~\ref{subsec:results-cid}).
The EEs are trained with a \emph{weighted-loss} function, similar to previous studies~\cite{kaya2019-sdn}. Further, we assign a \emph{higher weighting} factor, $w_{EV_k}$ for losses at EV-exits to minimize the effect of an overconfident $f_{out}$ (original output block). In addition, we increase each $w_{EV_k}$ by $\delta$ such that $w_{EV_k} = w_{EV_{k-1}} + \delta$ to vary the influence each EV-exit has on the weights of the shared base network, thereby promoting diversity. We empirically determine that $w_{EV_0}=3$ and $\delta=0.5$ yields the best results (see Appendix~\ref{subsec:apdx-train} for details). 
During training,

\vspace{-1mm}\noindent1)~Before each train batch, we copy weights of $k$\textsuperscript{th} EE-exit ($h_{\theta_k}$) into $k$\textsuperscript{th} EV-exit ($h_{\phi_k}$) i.e., $\mathbf{\boldsymbol{\phi_k = \theta_k}}$. Thus, each EV-exit iterates through a train batch with weights of the EE-exit. Weight transfer is limited solely from $h_{\theta_k}$ to $h_{\phi_k}$, with no other layers involved.

\vspace{-1mm}\noindent2)~After a train batch, the loss computed at an EV-exit is w.r.t. the weights copied from EE, after which weights of $h_{\phi_k}$ are updated during backpropagation. However, before the next train batch begins, we overwrite the weights of $h_{\phi_k}$ by copying from EE. Therefore, the weights learnt by layers before/after $h_{\phi_k}$ always align with EE weights. Thus, this does not affect convergence (see Figure~\ref{fig:convergence} and Section~\ref{subsec:results-ev-effectiveness-convergence}).
\begin{eqnarray}
     \mathcal{L} = \sum_{k=1}^K \mathcal{L}_{EE_k} +  \sum_{k=1}^K \mathcal{L}_{EV_k} + \mathcal{L}_{EF} \label{eq:train-obj}\\
     \begin{array}{cc}
         \mathcal{L}_{EE_k} = \tau_k \cdot \mathcal{L}_{CE} &  \quad \quad \quad \mathcal{L}_{EV_k} = w_{EV_k} \cdot \mathcal{L}_{CE} \nonumber
     \end{array}
\end{eqnarray}
The training objective reduces to minimizing Eqn~\ref{eq:train-obj}, where $\mathcal{L}$ is the total loss of \qute, which is the sum of the losses of EE, EV and final exits. $\mathcal{L}_{CE}$ is the cross-entropy loss, and $\mathcal{L}_{EF} = \mathcal{L}_{CE}$, which is loss of original final exit. $\mathcal{L}_{EE_k}$ and $\mathcal{L}_{EV_k}$ are losses at the k\textsuperscript{th} EE-exit and EV-exit respectively. $\tau_k$ is the weighting applied to EE loss~\cite{kaya2019-sdn}. $ w_{EV_k}$ is the weighting applied to the k\textsuperscript{th} EV-exit.

The weights of base network (only grey blocks in Figure~\ref{fig:qute}) are frozen for the last 10\% of epochs while continuing weight transfer, enabling isolated training of EV-exits to foster increased diversity. In this way, the EE-knowledge is transferred to the final exit(s) to obtain better uncertainty estimation on $\mathrm{S_{ID}}/\mathrm{S_{CID}}$. 
In contrast, without EV-assistance, the various final exits risk learning the same predictive distribution, leading to poor ensemble behavior. We compare the effectiveness of EV-assistance in Section~\ref{subsec:results-ev-effectiveness-convergence}.

\textbf{Inference}~During inference, the resource-heavy EE-blocks are removed from the model, leaving only the base network and the lighter output blocks $\{f_{out_k}(.)\}_{k=1}^{K}$. We also eliminate the original output block  $f_{out}$ due to its overconfident behavior, which compromises calibration quality. The final prediction ($p_{\Theta}(y|x)$) is obtained by calculating the mean of all $|\mathcal{K}|$ prediction vectors of $\mathcal{K}$.
\vspace{-1mm}
\begin{eqnarray}
    p_{\Theta}(y|x) = \frac{1}{|\mathcal{K}|}  \Bigl(\sum_{k=1}^{K} p_{{\phi}_k}(y|x)\Bigr) \label{eq:inference}
\end{eqnarray}

%% file: figs/ev-assist-horz.tex
\begin{tikzpicture}[font=\sffamily\small\fontfamily{LinuxBiolinumT-LF}\selectfont\Large,>=stealth']

  \tikzset{ionode/.style={font=\sffamily\small\fontfamily{LinuxBiolinumT-LF}\selectfont\large,>=stealth', align=center,minimum width=1.5cm,black!75}};
  \tikzset{layernode/.style={align=center,draw,minimum height=1cm, minimum width=1.5cm, rounded corners=3pt,inner sep=5pt,fill=black!5}};
  \tikzset{classfiernode/.style={align=center,draw,minimum height=3cm,rounded corners=4mm,inner xsep=8pt,, fill=yellow}};
  \tikzset{poolingnode/.style={align=center,draw,minimum height=2.5cm,rounded corners=5mm,inner xsep=8pt,fill=blue}};
  \tikzset{circlenode/.style={align=center,draw,minimum height=0.5cm,rounded corners=5mm,inner xsep=4pt,fill=blue}};
   \tikzset{eenode/.style={dotted, thick, align=center,draw, minimum height=0.7cm, minimum width=2.3cm, rounded corners=1mm,inner xsep=8pt,fill=red}};
   \tikzset{eenode_thick/.style={thick, align=center,draw, minimum height=0.7cm, minimum width=2.3cm, rounded corners=1mm,inner xsep=8pt,fill=red}};

  \node (anchor) [ionode,black] {};
  \node (fd) at (anchor.east) [layernode, black, xshift=1.5cm, yshift=-0.0cm, fill=black!10] {$f_{D}$};
  \node(ad_1) at (fd.west) [ionode, black, xshift=-0.7cm, font=\Large, yshift=0.3cm] {\textbf{$a_{D-1}$}};
  \node (houtk) at (fd.east) [layernode, black, xshift=1.5cm, yshift=0cm, fill=blue!10] {$h_{\phi_k}$};
  \node (softmax_fe) at (houtk.east) [eenode_thick,anchor=west, xshift=0.75cm, yshift=0cm, fill=yellow!30] {$\sigma(.)$};
  \node (pd_foutk) at (softmax_fe.east) [ionode, black, font=\Large, xshift=1.5cm, yshift=0cm] {\textbf{${p_{{\phi}_k}(y|x)}$}};
    
    \node (fi) at (fd.west) [layernode, black, xshift=-3.0cm, yshift=-0.0cm, fill=black!10] {$f_{i}$};
    \node(ai_1) at (fi.west) [ionode, black, xshift=-1cm, font=\Large, yshift=0cm] {\textbf{$a_{i-1}$}};

  \node (gk) at (fd.south) [eenode,anchor=west,xshift=-2cm, yshift=-1.5cm, fill=red!15] {$h_{{\theta}_{k}}$};
  \node (softmax_ee) at (gk.east) [eenode,anchor=west,xshift=1cm, yshift=0cm, fill=yellow!30] {$\sigma(.)$};
  \node (pd_gk) at (softmax_ee.east) [ionode, black, font=\Large, xshift=1.5cm, yshift=0.0cm] {\textbf{${p_{{\theta}_k}(y|x)}$}};

  \node (thetak) at (gk.east)  [ionode, black, font=\Large, xshift=-0.2cm, yshift=1.0cm] {$\theta_k$};
  \node (foutk_antte) at (houtk.west)  [ionode, black, font=\Large, xshift=4.5cm, yshift=-1.1cm] {$\mathbf{f_{out_k}}$};
  \node (gk_antte) at (gk.east)  [ionode, black, font=\Large, xshift=3.2cm, yshift=-0.8cm] {$\mathbf{g_{\theta_k}}$};

   \draw[->] (ai_1) -- (fi);
   \draw[->, dotted, thick] (fi.east)  -- (fd);
  \draw[->] (fd) -- (houtk);
  \draw[->] (houtk) -- (softmax_fe);
  \draw[->] (softmax_fe) -- (pd_foutk);
  \draw[->] (gk) -- (softmax_ee);
  \draw[->] (softmax_ee) -- (pd_gk);

  \draw[->] (fi.east) ++(0cm, 0.0cm) -- ++(0.5cm, 0) -- ++(0, -2cm) -- ++(0.5cm, 0);

  \draw[->, dotted, ultra thick] (gk) -- (houtk) ; 

  \draw[, dashed] (houtk.west) ++(-0.2, 0.75cm) -- ++(5.0cm, 0) -- ++(0, -1.5cm) -- ++(-5.0cm, 0) -- ++(0, 1.5cm);
  \draw[, dashed] (gk.west) ++(-0.2, 0.6cm) -- ++(6.0cm, 0) -- ++(0, -1.1cm) -- ++(-6.0cm, 0) -- ++(0, 1.1cm);

\end{tikzpicture}

%% file: tex/5-methodology.tex
\section{Evaluation Methodology}
\label{sec:methodology}

For our microcontroller (MCU) evaluations, we utilize two types of boards.

\vspace{-1mm}
1)~\bigmcu: The STM32 Nucleo-144 development board (STM32F767ZI)~\cite{stm32f767zi-datasheet} with 2MB flash and 512KB SRAM, clocked at 216MHz with a typical power consumption of 285mW. This board accommodates high-resource baselines that require substantial memory and processing power.

\vspace{-1mm}
2)~\smallmcu: The STM32 Nucleo-32 development board (STM32L432KC)~\cite{stm32l432kc-datasheet} with 256KB flash and 64KB RAM, clocked at 80MHz with a typical power consumption of 25mW. This ultra-low-power device better reflects our goal of deployment on smaller, energy-efficient systems. 


We evaluate \qute in three settings: 1) Accuracy-drop/CID detection, which monitors system performance over time to detect gradual shifts and temporal degradation patterns (Section~\ref{subsec:results-cid}), 2)~Failure detection, which targets instance-level identification of both misclassifications and OOD samples (Section~\ref{subsec:results-failure-detect}), and 3)~uncertainty estimation quality (Section~\ref{subsec:results-uq}). Since the focus of this work is to detect failures reliably in-the-field, accuracy-drop and failure detection experiments evaluate this capability.

\noindent\textbf{Datasets and Models}~We evaluate \qute on one audio classification and three image classification tasks of differing complexities: 1)~MNIST~\cite{mnist} on a 4-layer CNN, 2)~SpeechCommands~\cite{speech_commands} on a 4-layer depthwise-separable model (DSCNN)~\cite{kws} for keyword spotting task, 3)~CIFAR10~\cite{cifar-10} on Resnet-8 (from MLPerf tiny benchmark suite~\cite{mlperf-tiny-benchmark}) and 4)~TinyImagenet~\cite{le2015-tinyimagenet} on MobilenetV2~\cite{mobilenets}. While TinyImageNet/MobileNetV2 isn't a typical tinyML dataset/model, we include it to show \qute's broader applicability.

\vspace{-1mm}
\textbf{CID datasets}:~For CID datasets, we use corrupted versions of ID i.e., MNIST-C (15 corruptions)~\cite{mu2019-mnistc}, CIFAR10-C (19 corruptions) and TinyImagenet-C (15 corruptions)~\cite{hendrycks2019-benchmarking-corruptions}, with various corruptions such as frost, noise, fog, blur, pixelate etc.
For corrupting SpeechCmd, we use the audiomentations library~\cite{jordal2024-audiomentations}. We apply 11 types of corruptions such as noise, air absorption, time masking/stretching to obtain corrupted-SpeechCmd (denoted SpeechCmd-C).

\vspace{-1mm}
\textbf{OOD datasets}:~SpeechCommands has utterances of 35 words. Like~\citet{kws}, we train DSCNN to recognize ten words and use the rest as OOD (denoted SpeechCmd-OOD). Furthermore, we use FashionMNIST~\cite{xiao2017-fashionmnist} and SVHN~\cite{netzer2011reading-svhn} as OOD for MNIST and CIFAR10 respectively.
More details on models and datasets used can be found in Appendix~\ref{subsec:apdx-train}.

\vspace{-1mm}
\noindent\textbf{Placement of Early-exits}~The number of early-exits ($K$) is dictated by the resource budget. There are only a handful locations in a tinyML model where we can insert the early-exits. Our strategy is to insert them at equally-spaced locations along the base network. For MNIST and SpeechCmd, which both use 4-layer models, we insert two EEs after the 1\textsuperscript{st} and 2\textsuperscript{nd} layers. For Resnet-8  with 3 residual stacks, we insert EEs after 1\textsuperscript{st} and 2\textsuperscript{nd} residual stacks. We insert 5 EEs for MobilenetV2 at equally-spaced locations starting from the input. A detailed analysis of the effect of number of early-exits on UQ is provided in Appendix~\ref{subsec:apdx-ee-v-uncertainty}.

\vspace{-1mm}
\textbf{Accuracy-drop detection}~
Unlike with OOD data, correct predictions may still occur on CID inputs. Therefore, we focus on detecting accuracy-drop events caused by CID inputs. 
We present a realistic mechanism with no additional overhead that monitors \emph{confidence} to detect a possible drop in accuracy. We formulate this as a binary classification problem. First, for each baseline we evaluate, we iterate over \emph{only} ID and obtain model predictions while computing the moving average of accuracy of the past $m$ predictions using a sliding-window. The accuracy of the sliding-window is denoted as $\mathcal{A}_{SW}$ and the accuracy distribution thus obtained is denoted as $\mathcal{A}_{ID}$ with mean $\mu_{ID}$ and standard-deviation $\sigma_{ID}$ respectively. In our empirical evaluations, we find that $m=100$ is reasonable. 
We then construct corrupted datasets by appending ID with each corrupted-ID version.
Next, we iterate over these ID+CID datasets (one for each type of corruption)  using a sliding-window while computing the moving average of confidence ($\mathcal{C}_{SW}$). A \emph{CID-detected} event corresponds to $\mathcal{C}_{SW} < \rho$, and a true positive event occurs when  $\mathcal{C}_{SW} < \rho~ and ~\mathcal{A}_{SW} \leq \mu_{ID} - 3\cdot\sigma_{ID}$, where $\rho$ is a user-defined threshold. 
We vary $\rho$ from 0 to 1 in step sizes of 0.1 and repeat the whole experiment. Appendix~\ref{subsec:apdx-cid-results} provides additional details.
Finally, we report the average area under precision-recall curve (AUPRC). 

\vspace{-1mm}
\textbf{Failure detection}~Failure detection aims at rejecting both ID misclassifications and OOD instances, as both indicate a potential failure.
To this end, we define two binary classification tasks: 1)~ID\checkmark $|$ ID$\times$, which separates correct predictions from incorrect ones within ID/CID samples, and 2)~ID\checkmark $|$ OOD, which distinguishes correct predictions from OOD. For the former, we use ID+CID datasets where errors can arise from hard-to-classify ID samples or CID inputs, while for the latter, we use ID and OOD data.
In addition,~\cite{xia2023-failure-detection} leverages uncertainty estimation for failure detection, optimizing for a user-defined threshold. However, since this threshold may vary \textit{in the field} depending on application requirements, we report the threshold-independent Area Under the Receiver Operating Characteristic curve (AUROC). Finally,~\cite{xia2023-failure-detection} shows that Deep Ensembles (DEEP) achieve superior performance on threshold-free metrics, making DEEP the strongest baseline in our comparisons.

\vspace{-1mm}
\textbf{Uncertainty quantification}~
For UQ, we report the class-weighted F1 score instead of accuracy because its better for imbalanced datasets, and two proper scoring metrics~\cite{gneiting2007-proper-scoring-rules}: Brier score (BS)~\cite{brier1950verification}, which measures the accuracy of predicted probabilities and negative log-likelihood (NLL), which measures how close the predictions are to the ground truth (see Appendix~\ref{sec:apdx-metrics}). Also, despite its documented unreliability in measuring uncertainty quality~\cite{nixon2019-measuring-calib}, we report expected calibration error (ECE)~\cite{guo2017-ece} due to it popularity. 
Appendix~\ref{sec:apdx-metrics} discusses the limitations of ECE.

\vspace{1mm}
\textbf{BASELINES}~

\noindent\textbf{BASE}:~Unmodified implementations of models evaluated.

\vspace{-1mm}
\noindent\textbf{Monte Carlo Dropout (MCD)}~\cite{gal2016-MCdropout}:~computes mean of prediction vectors from $K$ inference passes by activating \emph{dropout}. For fairness, dropout and EE locations are same. We use a dropout rate of 0.1, 0.05, 0.2, 0.2  for MNIST, SpeechCmd, CIFAR10 and TinyImagenet respectively. 

\vspace{-1mm}
\noindent\textbf{Deep Ensembles (DEEP)}~\cite{lakshminarayanan2017-uncertainty-est-ensembles}:~explicit ensemble of models with same base model architecture but with different random weight initializations. The number of ensemble members = $K$.

\vspace{-1mm}
\noindent\textbf{Early-exit Ensembles (EE-ensemble)}~\cite{qendro2021-EEensemble}:~attaches multiple early-exits with additional FC-layers. The size of the FC layers is chosen such that the model delivers the best uncertainty estimation. All early-exit prediction vectors including that of the final exit are averaged to get the final ensemble prediction. We place the early-exits at the same locations as \qute. Ablation studies with EE-ensemble are presented in Appendix~\ref{subsec:apdx-ee-ensemble}.

\vspace{-1mm}
\noindent\textbf{HYDRA}~\cite{tran2020-hydra}:~uses the same architecture as \qute. Instead of early-exit distillation like \qute, it employs ensemble distillation, capturing unique predictive behavior of individual ensemble members. 

\vspace{-1mm}
\noindent\textbf{Generalized-ODIN (G-ODIN)}~\cite{hsu2020-generalized-odin}:~a generalized OOD detection method that decomposes the confidence score and introduces a preprocessing layer that perturbs the inputs for better OOD detection. The perturbation magnitude is determined on a small held-out validation set.

%% file: tex/6-results.tex
\section{Results}
\label{sec:results}

First, we evaluate our MCU implementations in Section~\ref{subsec:results-mcu} on both \smallmcu and \bigmcu to showcase \qute's resource efficiency and infeasibility of prior methods on ultra-low-power devices. \qute achieves an average \textbf{27\% latency reduction} on \bigmcu compared with the best performing baseline with \textbf{58\%} and \textbf{26\% smaller models} compared to EE-ensemble and DEEP respectively.  
On \smallmcu, high-resource prior methods \textbf{cannot even fit} on CIFAR10, further underscoring \qute's suitability for constrained environments. Next, we demonstrate \qute's superiority over all baselines in detecting accuracy-drop events due to CID in Section~\ref{subsec:results-cid}. On failure detection (Section~\ref{subsec:results-failure-detect}), \qute consistently outperforms prior methods in distinguishing between ID\checkmark $|$ ID$\times$ and is competitive on ID\checkmark $|$ OOD with other baselines, even outperforming specialized OOD detectors like G-ODIN on MNIST and SpeechCmd (Section~\ref{subsec:results-failure-detect}). Next, we compare \qute's calibration in Section~\ref{subsec:results-uq}, demonstrating that \qute predictions are consistently well-calibrated, especially on tiny-sized models with the \textbf{59\% lower model parameters} on average compared to EE-ensemble. In addition, we conduct an ablation study to assess the effectiveness of EV-assistance method and its effect on model convergence (Section~\ref{subsec:results-ev-effectiveness-convergence}). Finally, Sections~\ref{subsec:results-ts} and~\ref{subsec:results-postnet} demonstrate \qute's superiority in uncertainty estimation compared to temperature scaling (TS)~\cite{guo2017-ece, rahaman2021-uncertainty-quantification-ensembles} and single-pass deterministic methods~\cite{charpentier2020-postnets} respectively. Additional ablation studies are presented in Appendix~\ref{sec:apdx-ablation}.



\begin{figure*}[t]
    \centering
    \hspace{-1.5cm}
    \begin{subfigure}[]{0.2\textwidth}
        \centering
        \input{figs/mcu-results-size}
        \label{fig:mcu-size-big}
        \captionsetup{justification=centering, singlelinecheck=false}
        \caption{Size (KB)}        
    \end{subfigure}
    \hspace{-0.50cm}    
    \begin{subfigure}[]{0.2\textwidth}
        \centering
         \input{figs/mcu-results-latency}
        \label{fig:mcu-latency-big}
        \captionsetup{justification=centering, singlelinecheck=false}
        \caption{Latency (ms)}        
    \end{subfigure}
    \begin{subfigure}[]{0.2\textwidth}
        \centering
        \input{figs/mcu-results-size-small}
        \label{fig:mcu-size-small}
        \captionsetup{justification=centering, singlelinecheck=false}
        \caption{Size (KB)}        
    \end{subfigure}
    \hspace{-0.0cm}    
    \begin{subfigure}[]{0.2\textwidth}
        \centering
         \input{figs/mcu-results-latency-small}
        \label{fig:mcu-latency-small}
        \captionsetup{justification=centering, singlelinecheck=false}
        \caption{Latency (ms)}        
    \end{subfigure}
    \caption{Microcontroller results for SpeechCmd (sp-cmd), CIFAR10 (cfr10) on \bigmcu and \smallmcu (\textit{lower is better)}. EE-ensemble and DEEP on CIFAR10 \textbf{\textit{do not fit }} i.e., out-of-memory (OOM)}
    \label{fig:mcu-results-small}
\end{figure*}

\subsection{MCU Fit:~\qute vs Resource-Heavy methods}
\label{subsec:results-mcu}
We evaluate MCU implementations of \qute on two MCUs of different sizes. Since the primary objective of TinyML systems is to minimize the \textit{energy-per-prediction}~\cite{ghanathe2023trecx}, our target platform is the \smallmcu, as the base network fits comfortably on this smaller device. 


We compare \qute against the two most relevant baselines: EE-ensemble and DEEP. We exclude MCD due to its reliance on a specialized dropout module~\cite{ahmed2023-spindrop-droput-module}, which is impractical on MCUs. We also exclude HYDRA because it is often suboptimal, as shown in the following sections.
Our evaluation uses CIFAR10 and SpeechCommands, as TinyImageNet/MobileNetV2 exceeds tinyML device limits.  On \bigmcu, \qute achieves \textbf{31\% and 47\% latency reductions} over EE-ensemble and DEEP, respectively, and maintains accuracy parity with both, even with \textbf{58\% and 26\% smaller models}. 
On \smallmcu, both EE-ensemble and DEEP \textbf{\textit{do not fit}} on the device for CIFAR10; EE-ensemble model size exceeds \smallmcu's on-device memory by 270KB, whereas DEEP exceeds by 1.9KB. 
Notably, EE-ensemble has the highest peak RAM usage than DEEP and \qute due to its need to store large intermediate feature maps during early-exit computation. In contrast, DEEP processes each ensemble member sequentially as there is no parallel compute capability on MCUs. This characteristic allows single-forward-pass methods like \qute and EE-ensemble to achieve reduced latency, especially as the ensemble size increases. Appendix~\ref{subsec:apdx-mcu} delves into more details.
These results underscore that the novelty of \qute lies in its exceptional resource efficiency, making it a practical solution for deploying UQ or model monitoring mechanisms on ultra-low-power MCUs, especially in scenarios where other approaches may be infeasible. This is crucial for the extreme edge because sometimes the alternative is no UQ/monitoring.

\input{figs/table-qute-aleatoric-epistemic}

\begin{figure*}[t]
    \centering
    \begin{subfigure}[]{0.3\textwidth}
            \input{figs/table-cid-mnist-speechcommands}
        \caption{AUPRC MNIST-C \& SpeechCmd-C}
                \label{table:cid-detect-mnist-speechcmd-auprc}
        \hspace{-0.5cm}
    \end{subfigure}
    \begin{subfigure}[]{0.3\textwidth}
                \centering
                \scalebox{0.8}{
                \input{figs/cid-detect-cifar10-auprc}
                }
        \caption{AUPRC CIFAR10-C}
        \label{fig:cid-detect-cifar10-auprc}
        \end{subfigure}
                \hspace{0cm}
        \begin{subfigure}[]{0.3\textwidth}
                \centering
                \scalebox{0.8}{
                \input{figs/cid-detect-timgnet-auprc}
                }
        \caption{AUPRC TinyImagenet-C}
         \label{fig:cid-detect-timgnet-auprc}
    \end{subfigure}
    \caption{Accuracy-drop detection results. AUPRC is reported for all evaluated baselines (\emph{higher is better})}
    \label{fig:cid-detects}
\end{figure*}

    


\subsection{Accuracy-drop detection}
\label{subsec:results-cid}


As described in Section~\ref{sec:methodology}, we monitor for a drop in confidence to detect accuracy-drop events in a model due to CID. We also explore alternative uncertainty measures like predictive entropy. However, we choose confidence because it produces similar results without extra processing, a practical advantage for tinyML. Figure~\ref{fig:cid-detects} reports AUPRC averaged over all corruptions for all datasets. 
\qute consistently outperforms all baselines in detecting accuracy-drops caused due to CID across all datasets, which is crucial for maintaining model reliability in deployment. Our empirical evaluations show that early-exits (EE) are more adept at capturing  epistemic uncertainty whereas deeper layers are better at capturing aleatoric uncertainty, as illustrated by Table~\ref{table:qute_uncertainty}. Table~\ref{table:qute_uncertainty} reports the percentage reduction in average prediction confidence of models of different sizes on fog (epistemic) and gaussian noise (aleatoric) corruptions compared to the average confidence on CIFAR10-ID.
By combining knowledge from EE at the final exits, \qute exhibits reduced confidence on both types of corruptions, thereby enhancing its accuracy-drop detection capabilities.

MCD performs poorly in all cases either due to reduced model capacity (due to dropout) or due to extreme calibration. For example, on TinyImagenet-ID, MCD with an accuracy of 0.34 has a median confidence value of 0.345 across all predictions. However, on TinyImagenet-CID, the overparameterized nature of modern neural networks does not allow the median confidence of MCD to drop any further than 0.27 i.e., a drop of 0.075, which is not discernible enough to detect drop in accuracy. This is observed with EE-ensemble too for larger model sizes. In contrast, \qute's median confidence on TinyImagenet-ID is 0.69 and the average median confidence on CID is 0.57, which helps create a much clearer distinction between ID and CID. EE-ensemble performs comparably to \qute for tiny/medium-sized models. 
However, for large models, the additional learning layers added to balance the learning capacities of all exits cause the early-exits to function like conventional deep networks. Therefore, EE-ensemble becomes overconfident for severe corruptions, which results in a 8.9\% drop in performance on TinyImagenet-C. In contrast, the early-view-assistance method allows \qute to retain early-exit knowledge in its output blocks across different model sizes, enabling it to outperform all baselines on CID detection on all datasets/models. The performance of \qute and EE-ensemble (especially on tiny-sized models), highlight the utility of early-exit knowledge in detecting corruptions. 
Similarly, G-ODIN remains overconfident on CID, and is surpassed by most baselines on all corrupted datasets. This suggests that general OOD-detectors are not automatically effective for CID detection, consistent with previous findings that OOD detectors find it hard to detect non-semantic shifts~\cite{hsu2020-generalized-odin}. Further, HYDRA is consistently subpar at accuracy-drop detection, often being outperformed even by BASE. This highlights HYDRA's need for larger classification heads with more parameters to fully integrate ensemble-knowledge, consistent with the original paper's methodology. Contrarily, \qute's early-exit distillation method proves effective in detecting corruptions compared to ensemble distillation mechanisms.

\subsection{Failure detection}
\label{subsec:results-failure-detect}
\input{figs/table-failure-detection}

Table~\ref{table:failure-detection} reports the AUROC for failure detection experiments for ID\checkmark $|$ ID$\times$ and ID\checkmark $|$ OOD. As seen, for ID\checkmark $|$ ID$\times$, \qute outperforms all baselines on MNIST and SpeechCmd. MCD performs slightly better on CIFAR10, likely due to the higher number of epistemic uncertainty-inducing corruptions present in CIFAR10-C, such as blur and digital distortions. This observation aligns with prior work highlighting MCD's strength in capturing instance-level epistemic uncertainties~\cite{kendall2017-unceratainties-bayesian-DL}. In contrast, \qute effectively captures both epistemic and aleatoric uncertainties, resulting in overall superior performance across datasets.
For ID\checkmark $|$ OOD, \qute outperforms all baselines on SpeechCmd, and is a close second on MNIST. Surprisingly, G-ODIN, a specialized OOD detector, is poor on MNIST and SpeechCmd, even outperformed by BASE. 
Further analysis revealed that G-ODIN is under-confident on these ID-datasets and over-confident on OOD. As an ablation study, we increased the number of training epochs of G-ODIN by 30 epoch, which resulted a 55\% improvement on ID\checkmark $|$ OOD on MNIST. Contrarily, on a larger model, Resnet-8 on CIFAR10, G-ODIN's performance notably improves on ID\checkmark $|$ OOD, consistent with previous findings. This may suggest the need to rethink OOD detection with extremely tiny models. These results showcase \qute's versatility/efficacy in detecting both CID and OOD data.

\subsection{Uncertainty quantification}
\label{subsec:results-uq}
Table~\ref{table:uq} reports the calibration metrics for all datasets we evaluate. \qute \emph{outperforms} all baselines on tiny-sized models i.e., MNIST and SpeechCmd, and is \emph{competitive} for medium and large models i.e., CIFAR10 and TinyImagenet despite using significantly lower resources. 
All methods perform better than BASE, except MCD on MNIST and SpeechCmd. Further inquiry revealed that applying dropout during inference on the extremely tiny models takes away from its model capacity thereby leading to drop in accuracy and poor calibration. 
DEEP showcases good calibration on all datasets owing to its larger model capacity. However, both MCD and DEEP are impractical for tinyML because they both require multiple inference passes or specialized hardware.  
In contrast, \qute provides comparable uncertainty estimates in a single forward pass with 53\% lower model sizes on average compared to DEEP. 

\qute also outperforms EE-ensemble, the most relevant prior work, across all datasets with models that are, on average, 59\% smaller, except on larger datasets and models like TinyImagenet on MobileNetV2. Since \qute's primary focus is resource efficiency, the limited parameters in its output blocks—restricted to a single depthwise convolutional (CONV) layer—can constrain calibration performance relative to high-resource methods like DEEP and EE-ensemble. To normalize the comparison, we enhanced the \qute architecture with additional learning layers at each output block, creating \qute$+$. For Resnet-8 on CIFAR10, we added two depthwise-separable CONV layers, while for MobileNetV2 on TinyImageNet, we included an additional dense (fully connected) layer to form \qute$+$. As seen in Table~\ref{table:uq}, F1 score of \qute$+$ exceeds that of DEEP on CIFAR10, however, it slightly degrades on TinyImagenet. Further analysis revealed that since the base network is trained simultaneously with all early-exits (EE) and EV-exits, the weights of the shared base network is negatively impacted since the training routine tries to optimize for all exits. In contrast, EE-ensemble method employs only half as many exits (only EEs) compared to \qute, and the EE's regularization effect helps network accuracy~\cite{branchynet}. Appendix~\ref{subsec:apdx-ee-v-uncertainty} studies the effect of ensemble size on \qute's calibration performance. Nevertheless, \qute$+$ significantly improves calibration, the primary focus of this work, outperforming all baselines on BS and achieving NLL on par with the top-performing method. These results indicate that, in an unconstrained setting, \qute can achieve calibration performance comparable to high-resource methods. Additionally, we assess the effectiveness of \qute's early-exit distillation technique by comparing against HYDRA, which utilizes ensemble-distillation. While HYDRA’s lightweight classification heads (similar to \qute) capture ensemble knowledge in tiny models, they struggle as model size increases, leading to reduced accuracy and poor calibration, even underperforming BASE on CIFAR10. This reveals HYDRA’s reliance on larger classification heads for effective ensemble integration as dicussed before. Contrarily, \qute's early-exit distillation method consistently proves effective across across diverse models and datasets.

\input{figs/table-uq-new}

\input{figs/table-ev-effectiveness}
\subsection{Effectiveness of EV-assistance method and its effect on convergence}
\label{subsec:results-ev-effectiveness-convergence}
\begin{figure}
    \centering
    \vspace{0.0cm}
    \scalebox{0.25}{
        \includegraphics[width=\textwidth-0.2cm]{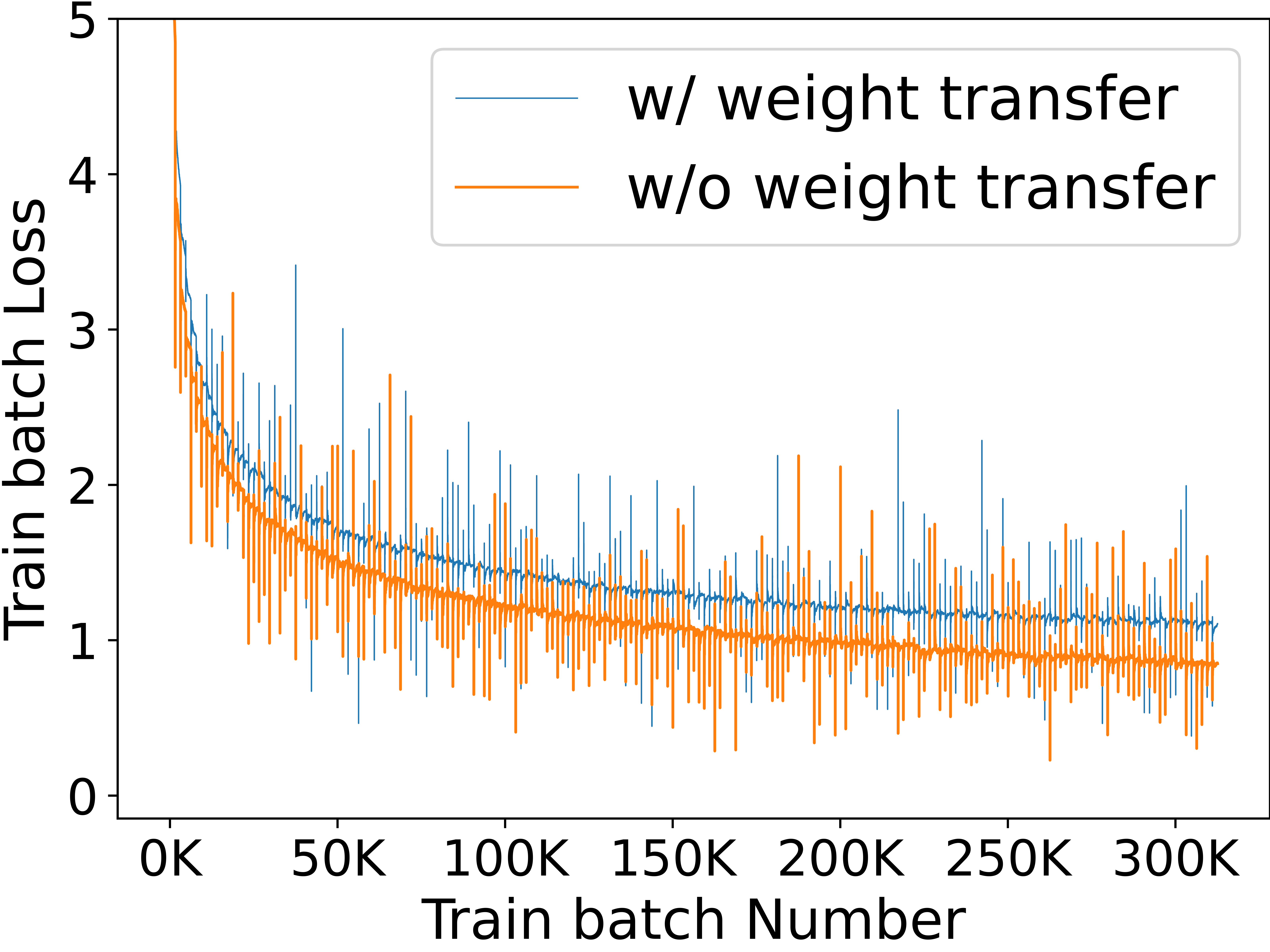}
    }
    \captionsetup{justification=justified}
    \caption{Batch-loss at EV-0 for CIFAR10 on Resnet-8\vspace{0cm}}
    \label{fig:convergence}
\end{figure}
To investigate the effect of the early-view (EV) assistance method on ensemble quality, we train non-EV versions of \qute i.e., the same architecture (with early-exits) except without the weight transfer mechanism. Table~\ref{table:ev-effectiveness} reports the calibration metrics for both EV and non-EV versions for CIFAR10 and TinyImagenet. As shown, EV-assistance has a massive influence on improving calibration, and improves the network accuracy on TinyImagenet by 8\%. Also, we find that the ability of non-EV versions to detect accuracy-drop events reduces by 15\% on average compared to EV-versions. 
These results show that the notably lower NLL obtained with EV-assistance (5\% lower on CIFAR10 and 19\% lower on TinyImagenet) not only improves uncertainty estimates, but it also proves crucial in both CID and OOD detection. In addition, we investigate the effect of weight transfer on model convergence. Figure~\ref{fig:convergence} plots the loss after each train-batch at one of the EV-blocks of Resnet-8 on CIFAR10. As seen, model convergence is not affected because the weight transfer mechanism is designed such that the loss at EV-exit is always computed w.r.t. to the copied weights (see Section~\ref{sec:design}). In essence, EV-assistance uses the EE-weights to extract and reemphasize the EE features at the final exit(s).

\input{figs/table-TS}
\input{figs/table-cid-detect-ts}

\subsection{Comparison with Temperature scaling}
\label{subsec:results-ts}

Temperature scaling (TS)~\cite{guo2017-ece} is a simple post-training calibration technique that applies a linear scaling factor to logits (pre-softmax) to mitigate model overconfidence. Table~\ref{table:uq-ts} and Table~\ref{table:cid-ts} present the calibration and accuracy-drop detection results, respectively, for TS versions of BASE (BASE-TS) and \qute (\qute-TS). For \qute-TS, the same architecture as \qute is used, but without weight transfer. We employed the best-performing \textit{pool-then-calibrate} configuration from~\cite{rahaman2021-uncertainty-quantification-ensembles}, which applies temperature scaling to ensembles to obtain \qute-TS. As the results show, \qute consistently outperforms BASE-TS on both calibration and accuracy-drop detection. However, despite improving calibration in some scenarios, \qute-TS performs worse than \qute on accuracy-drop detection, particularly with tiny datasets such as MNIST-C and SpeechCmd-C. A deeper analysis revealed that \qute-TS tends to be overconfident for certain corruptions. For instance, under fog corruption on MNIST-C, \qute-TS exhibits an overconfident behavior with a median confidence of 0.99, while \qute maintains a much lower confidence of 0.51, leading to superior accuracy-drop detection capability.

\subsection{Comparison with single-pass deterministic methods}
\label{subsec:results-postnet}
\input{figs/table-uq-postnet}

Postnets (PostN)~\cite{charpentier2020-postnets} is the most relevant single-pass deterministic method compared to \qute, which uses normalizing flows to predict posterior distribution of any input sample with no additional memory overhead. We evaluate PostN on MNIST and CIFAR10 by substituting the encoder network architectures with the ones used in our evaluations. We use identical hyperparameter settings of~\cite{charpentier2020-postnets} and train for same number of epochs as in our evaluations but with early stopping. Table~\ref{table:uq-postn} reports the calibration metrics for PostN and \qute. Since PostN don't necessarily focus on accuracy, we found that MNIST-PostN needed a 50\% increase in number of epochs to even surpass the accuracy of MNIST-BASE. We report this result. For CIFAR10, we found that early stopping halted the training, and report those results. As seen, \qute consistently outperforms PostN on all metrics on both datasets, making \qute a more efficient choice for resource-constrained environments.


%% file: figs/mcu-results-size.tex
\begin{tikzpicture}
	\pgfplotstableread{
        x   xlabel		   base	    ee-ensemble	    deep        qute   
        1    sp-cmd       294.5      361.5          321.6       297.8
        3    cfr10		 344.4          735.5       423.4       355.3

	}\data
		\begin{axis}[
			width=\textwidth+ 0.0cm,
			height=\graphHeight*2 - 3.5cm ,
			%
			major tick length=3pt,
			%
			major grid style={dashed,color=gray!50},
			minor grid style={color=gray!50},
			ymajorgrids=true,
			xmajorgrids=true,
			%
			xtick=data,
			xticklabels from table={\data}{xlabel},
			xticklabel style={text height=5pt,font=\scriptsize, rotate=0},
			%
			xlabel style={font=\scriptsize},
		    xlabel near ticks,
                xmin = 0.5, xmax = 3.5,
		   extra x ticks={ 0.5, 1.5, 2.5, 3.5},
			extra x tick style={
				grid=minor,
		    	xticklabel=\empty
			},
			%
			ymin=0,ymax=750,
			yticklabel style={font=\scriptsize},
			%
			ylabel style={align=center,font=\scriptsize},
               legend style={at={(0.55,1.0)},anchor=south, draw=none, cells={align=left}, font=\scriptsize,text width=2.3cm},
			legend columns=1,
			%
			ybar=3pt,
			area legend,
			bar width=3pt,
			enlarge x limits=0.1,
			%
		]



  \addplot [bar2] table[x=x, y=base, meta=base] {\data};
        \addplot [bar3]  table[x=x, y=ee-ensemble, meta=ee-ensemble] {\data};
        \addplot [bar4]  table[x=x, y=deep, meta=deep] {\data};
        \addplot [bar1] table[x=x, y=qute, meta=qute] {\data};

        
		
		


        
		\end{axis}
    \node at (0.2, 2) [anchor=west] {\textbf{\tiny \bigmcu}};

	\end{tikzpicture}

%% file: figs/mcu-results-latency.tex
\begin{tikzpicture}
	\pgfplotstableread{
        x   xlabel		   base	    ee-ensemble	   deep      qute
        1    sp-cmd       22.7     51.7    45.407      24.07	
        3    cfr10		 58.04   65.44     116.283     59.6
             
	}\data
		\begin{axis}[
			width=\textwidth+0cm,
			height=\graphHeight*2 - 3.5cm ,
			%
			major tick length=3pt,
			%
			major grid style={dashed,color=gray!50},
			minor grid style={color=gray!50},
			ymajorgrids=true,
			xmajorgrids=true,
			%
			xtick=data,
			xticklabels from table={\data}{xlabel},
			xticklabel style={text height=5pt,font=\scriptsize, rotate=0},
			%
			xlabel style={font=\scriptsize},
		    xlabel near ticks,
                xmin = 0.5, xmax = 3.5,
		   extra x ticks={ 0.5, 1.5, 2.5, 3.5},
			extra x tick style={
				grid=minor,
		    	xticklabel=\empty
			},
			%
			ymin=0,ymax=120,
			yticklabel style={font=\scriptsize},
			%
			ylabel style={align=center,font=\scriptsize},
               legend style={at={(2.32,0.5)},anchor=east, draw=none, cells={align=left}, font=\tiny,text width=1cm,},
			legend columns=1,
			%
			ybar=3pt,
			area legend,
			bar width=3pt,
			enlarge x limits=0.1,
			%
		]



  \addplot [bar2] table[x=x, y=base, meta=base] {\data};
        \addplot [bar3]  table[x=x, y=ee-ensemble, meta=ee-ensemble] {\data};
        \addplot [bar4]  table[x=x, y=deep, meta=deep] {\data};
        \addplot [bar1] table[x=x, y=qute, meta=qute] {\data};

		
		


        
		\end{axis}
        \node at (0.2, 2) [anchor=west] {\textbf{\tiny \bigmcu}};
	\end{tikzpicture}

%% file: figs/mcu-results-size-small.tex
\begin{tikzpicture}
	\pgfplotstableread{
        x   xlabel		   base	    ee-ensemble	    deep    qute
        1    sp-cmd       84.8     157.7            111.3     93  
        3    cfr10		 187.5   0   0       201

	}\data
		\begin{axis}[
			width=\textwidth+ 0.0cm,
			height=\graphHeight*2 - 3.5cm ,
			%
			major tick length=3pt,
			%
			major grid style={dashed,color=gray!50},
			minor grid style={color=gray!50},
			ymajorgrids=true,
			xmajorgrids=true,
			%
			xtick=data,
			xticklabels from table={\data}{xlabel},
			xticklabel style={text height=5pt,font=\scriptsize, rotate=0},
			%
			xlabel style={font=\scriptsize},
		    xlabel near ticks,
                xmin = 0.5, xmax = 3.5,
		   extra x ticks={ 0.5, 1.5, 2.5, 3.5},
			extra x tick style={
				grid=minor,
		    	xticklabel=\empty
			},
			%
			ymin=0,ymax=220,
			yticklabel style={font=\scriptsize},
			%
			ylabel style={align=center,font=\scriptsize},
               legend style={at={(0.55,1.0)},anchor=south, draw=none, cells={align=left}, font=\scriptsize,text width=2.3cm},
			legend columns=1,
			%
			ybar=3pt,
			area legend,
			bar width=3pt,
			enlarge x limits=0.1,
			%
		]



  \addplot [bar2] table[x=x, y=base,] {\data};
        \addplot [bar3]  table[x=x, y=ee-ensemble, ] {\data};
        \addplot [bar4]  table[x=x, y=deep] {\data};
        \addplot [bar1] table[x=x, y=qute, ] {\data};

        \node at (axis cs:2.8, 60)  [rotate=90]{\tiny \texttt{\textbf{OOM}}}; 
        \node at (axis cs:3.2, 60)  [rotate=90] {\tiny \texttt{\textbf{OOM}}}; 
		
		


        
		\end{axis}
        \node at (0.2, 2) [anchor=west] {\textbf{\tiny \smallmcu}};
	\end{tikzpicture}

%% file: figs/mcu-results-latency-small.tex
\begin{tikzpicture}
	\pgfplotstableread{
        x   xlabel		   base	    ee-ensemble	   deep      qute
        1    sp-cmd       160.706   376.387    321.221      173.136	
        3    cfr10		 291.266    0           0        298.056
             
	}\data
		\begin{axis}[
			width=\textwidth+0cm,
			height=\graphHeight*2 - 3.5cm ,
			%
			major tick length=3pt,
			%
			major grid style={dashed,color=gray!50},
			minor grid style={color=gray!50},
			ymajorgrids=true,
			xmajorgrids=true,
			%
			xtick=data,
			xticklabels from table={\data}{xlabel},
			xticklabel style={text height=5pt,font=\scriptsize, rotate=0},
			%
			xlabel style={font=\scriptsize},
		    xlabel near ticks,
                xmin = 0.5, xmax = 3.5,
		   extra x ticks={ 0.5, 1.5, 2.5, 3.5},
			extra x tick style={
				grid=minor,
		    	xticklabel=\empty
			},
			%
			ymin=0,ymax=400,
			yticklabel style={font=\scriptsize},
			%
			ylabel style={align=center,font=\scriptsize},
               legend style={at={(2.32,0.5)},anchor=east, draw=none, cells={align=left}, font=\tiny,text width=1cm,},
			legend columns=1,
			%
			ybar=3pt,
			area legend,
			bar width=3pt,
			enlarge x limits=0.1,
			%
		]



  \addplot [bar2] table[x=x, y=base, meta=base] {\data};
        \addplot [bar3]  table[x=x, y=ee-ensemble, meta=ee-ensemble] {\data};
        \addplot [bar4]  table[x=x, y=deep, meta=deep] {\data};
        
        \addplot [bar1] table[x=x, y=qute, meta=qute] {\data};

		
		
		\addlegendentry{BASE}
              \addlegendentry{EE}
              \addlegendentry{DEEP}
              \addlegendentry{\qute}
    
        
        \node at (axis cs:2.8, 100)  [rotate=90]{\tiny \texttt{\textbf{OOM}}}; 
        \node at (axis cs:3.2, 100)  [rotate=90] {\tiny \texttt{\textbf{OOM}}}; 
        
        
		\end{axis}
        \node at (0.2, 2) [anchor=west] {\textbf{\tiny \smallmcu}};
	\end{tikzpicture}

%% file: figs/table-qute-aleatoric-epistemic.tex
\begin{table}[t]
    \centering
    \scalebox{0.7}{
    \begin{tabular}{@{}lccccc@{}}
        \toprule
        \textbf{Model}      & \textbf{1stack} & \textbf{2stack} & \textbf{3stack} & \textbf{4stack} & \textbf{\qute}\\ \midrule
        \textbf{Fog}        & 30\%             & 24\%            & 18\%            & 11\%        & 25\%\\
        \textbf{Gaussian Noise} & 7\%          & -1.2\%         & 12\%            & 13\%    & 20\%        \\
        
    \end{tabular}
    }
    \caption{Percentage reduction in average confidence for Fog (high epistemic) and Gaussian Noise (high aleatoric) corruptions in Resnet\{1,2,3,4\} and Resnet- \qute.}
    \label{table:qute_uncertainty}
\end{table}

%% file: figs/table-cid-mnist-speechcommands.tex
\scalebox{0.7}{

{

\begin{tabular}{l | c | c }
 \textbf{AUPRC} ($\uparrow$)&   MNIST-C & SpeechCmd-C\\
\hline

\hline
~-~BASE & 0.54  &	0.52  \\ 	

~-~MCD & 0.45  &	0.56  \\ 	

~-~DEEP & 0.55  &	0.57  \\ 	

~-~EE-Ensemble & 0.63  & 0.58 \\	

~-~G-ODIN & 0.48  &	0.37  \\ 	

~-~HYDRA & 0.53  &	0.51  \\ 	

~-~\qute & \textbf{0.63}  &	\textbf{0.61}  \\ 	

\hline

\end{tabular}
}
}

%% file: figs/cid-detect-cifar10-auprc.tex
\begin{tikzpicture}
	\pgfplotstableread{
        sev		base	mcd	  deep	   ee-ensemble	godin	hydra  qute
        1		0.29	0.23    0.26	0.24	   0.19	   0.22   0.30
        2		0.39	0.31	0.41	0.38	   0.32	   0.28   0.47
        3		0.38	0.4 	0.53	0.51       0.37	   0.26   0.52
        4		0.49	0.48    0.63	0.64       0.50	   0.29   0.65
        5		0.62	0.72	0.72	0.75	   0.59	   0.45   0.78
             
	}\data
		\begin{axis}[
			width=\textwidth,
			height=\graphHeight*2 - 2.6cm ,
			%
			major tick length=3pt,
			%
			major grid style={dashed,color=gray!50},
			minor grid style={color=gray!50},
			ymajorgrids=true,
			xmajorgrids=true,
			%
			xticklabel style={text height=5pt,font=\scriptsize},
			%
			xlabel = Severity,
			xlabel style={font=\scriptsize},
		    xlabel near ticks,
                xmin = 1,
                xmax = 5,
		   extra x ticks={1, 2 ,...,5},
			%
			ymin=0.15,ymax=0.85,
			yticklabel style={font=\scriptsize},
			%
			ylabel style={align=center,font=\scriptsize},
               legend style={at={(0.55,1.0)},anchor=south, draw=none, cells={align=left}, font=\scriptsize,text width=2.3cm},
			legend columns=1,
			%
			enlarge x limits=0.1,
			%
		]



  \addplot+[rred, smooth, mark=o] table[x=sev, y=base] {\data};
        \addplot+[orange, smooth, mark=star] table[x=sev, y=mcd] {\data};
        \addplot+[ggreen, smooth, mark=square*] table[x=sev, y=deep] {\data};
        \addplot+[black, smooth,  mark=triangle*] table[x=sev, y=ee-ensemble] {\data};
        \addplot+[brown, smooth, mark=diamond*] table[x=sev, y=godin] {\data};
        \addplot+[yyellow, smooth, mark=pentagon*] table[x=sev, y=hydra] {\data};
        \addplot+[bblue, smooth, ultra thick, mark=*] table[x=sev, y=qute] {\data};

		
		


        
		\end{axis}
	\end{tikzpicture}

%% file: figs/cid-detect-timgnet-auprc.tex
\begin{tikzpicture}
	\pgfplotstableread{
	   sev		base	mcd	  deep	ee-ensemble	godin	hydra  qute
            1		0.14	0.08	0.15	0.17	0.32	0.11  0.2
            2		0.3	    0.08	0.5	    0.53	0.59	0.41  0.53
            3		0.47	0.17	0.73	0.67	0.67	0.52  0.76
            4		0.52	0.4	    0.73	0.61	0.69	0.63  0.77
            5		0.59	0.51	0.77	0.67	0.74	0.58  0.82

	}\data
		\begin{axis}[
			width=\textwidth,
			height=\graphHeight*2 - 2.6cm ,
			%
			major tick length=3pt,
			%
			major grid style={dashed,color=gray!50},
			minor grid style={color=gray!50},
			ymajorgrids=true,
			xmajorgrids=true,
			%
			xticklabel style={text height=5pt,font=\scriptsize},
			%
			xlabel = Severity,
			xlabel style={font=\scriptsize},
		    xlabel near ticks,
                xmin = 1,
                xmax = 5,
		   extra x ticks={1, 2 ,...,5},
			%
			ymin=0,ymax=0.9,
			yticklabel style={font=\scriptsize},
			%
			ylabel style={align=center,font=\scriptsize},
               legend style={at={(1.7,0.5)},anchor=east, draw=none, cells={align=left}, font=\scriptsize,text width=0.9cm},
			legend columns=1,
			%
			enlarge x limits=0.1,
			%
		]

		  \addplot+[rred, smooth, mark=o] table[x=sev, y=base] {\data};
        \addplot+[orange, smooth, mark=star] table[x=sev, y=mcd] {\data};
        \addplot+[ggreen, smooth, mark=square*] table[x=sev, y=deep] {\data};
        \addplot+[black, smooth,  mark=triangle*] table[x=sev, y=ee-ensemble] {\data};
        \addplot+[brown, smooth, mark=diamond*] table[x=sev, y=godin] {\data};
        \addplot+[yyellow, smooth, mark=pentagon*] table[x=sev, y=hydra] {\data};
        \addplot+[bblue, smooth, ultra thick, mark=*] table[x=sev, y=qute] {\data};

		
		
		\addlegendentry{BASE}
              \addlegendentry{MCD}
              \addlegendentry{DEEP}
              \addlegendentry{EE}
              \addlegendentry{G-ODIN}
              \addlegendentry{HYDRA}
              \addlegendentry{\qute}


        
		\end{axis}
	\end{tikzpicture}

%% file: figs/table-failure-detection.tex
\def\checkmark{\tikz\fill[scale=0.3](0,.35) -- (.25,0) -- (1,.7) -- (.25,.15) -- cycle;}

\begin{table}[t]
\centering
\small
\scalebox{0.8}{

{

\begin{tabular}{l | c  c  c | c  c  c }
 \textbf{AUROC} & \multicolumn{3}{c|}{ID\checkmark $|$ ID$\times$} & \multicolumn{3}{c}{ID\checkmark $|$ OOD}\\
 
 & Mnist & SpCmd & cfr10 & Mnist & SpCmd & cfr10\\
 
\hline

\hline

~-BASE        & 0.75         & 0.90         & 0.84         & 0.07         & 0.9          & 0.88         \\
~-MCD         & 0.74         & 0.89         & \textbf{0.87} & 0.48         & 0.89         & 0.89         \\
~-DEEP        & 0.85         & 0.91         & 0.86         & 0.78         & 0.91         & \underline{0.92}         \\
~-EE-ensemble & 0.85         & 0.90         & 0.85         & \textbf{0.85} & 0.90         & 0.90         \\
~-G-ODIN      & 0.72         & 0.74         & 0.83         & 0.4          & 0.74         & \textbf{0.95} \\
~-HYDRA       & 0.81         & 0.90         & 0.83         & 0.71         & 0.9          & 0.90         \\
~-\qute        & \textbf{0.87} & \textbf{0.91} & \underline{0.86}         & \underline{0.84}         & \textbf{0.91} & 0.91         \\


\end{tabular}
}
}
\caption{Failure detection experiments}
\label{table:failure-detection}
\end{table}



 











%% file: figs/table-uq-new.tex
\begin{table}[]
\centering
\small
\scalebox{0.8}{
{
\begin{tabular}{l | c c c c }
\hline
\textbf{Model}  & \textbf{F1 ($\uparrow$)}  & \textbf{BS ($\downarrow$)} & \textbf{NLL ($\downarrow$)} & \textbf{ECE ($\downarrow$)} \\
\hline

\textbf{MNIST} & \multicolumn{4}{c}{} \\
 ~-BASE &   0.910$\pm$0.002 & 0.013$\pm$0.000 & 0.292$\pm$0.006 & 0.014$\pm$0.001    \\
 
 ~-MCD      & 0.886$\pm$0.004 & 0.018$\pm$0.000 & 0.382$\pm$0.004 & 0.071$\pm$0.006   \\
 
 ~-DEEP   &      0.931$\pm$0.005 & 0.010$\pm$0.000 & 0.227$\pm$0.002 & 0.034$\pm$0.004    \\
 
 ~-EE-ensemble  &       0.939$\pm$0.002 & 0.011$\pm$0.000 & 0.266$\pm$0.005 & 0.108$\pm$0.002  \\

  ~-HYDRA  &     0.932$\pm$0.006 & 0.010$\pm$0.000 & 0.230$\pm$0.012 & \textbf{0.014$\pm$0.005}    \\

 ~-\qute &  \textbf{0.941$\pm$0.004} & \textbf{0.009$\pm$0.000} & \textbf{0.199$\pm$0.010} & 0.026$\pm$0.003   \\

\hline

\textbf{SpeechCmd}  & \multicolumn{4}{c}{} \\   
 ~-BASE &    0.923$\pm$0.007 & 0.010$\pm$0.000 & 0.233$\pm$0.016 & 0.026$\pm$0.001         \\
 
 ~-MCD &     0.917$\pm$0.006 & 0.011$\pm$0.000 & 0.279$\pm$0.013 & 0.048$\pm$0.002         \\
 
 ~-DEEP &      \textbf{0.934$\pm$0.008} & 0.008$\pm$0.000 & 0.205$\pm$0.012 & 0.034$\pm$0.006        \\
 
 ~-EE-ensemble     & 0.926$\pm$0.002 & 0.009$\pm$0.000 & 0.226$\pm$0.009 & 0.029$\pm$0.001       \\
 
 ~-HYDRA &       0.932$\pm$0.005 & 0.008$\pm$0.000 & 0.203$\pm$0.016 & 0.018$\pm$0.004          \\
 
 ~-\qute &         0.933$\pm$0.006 & \textbf{0.008$\pm$0.000} & \textbf{0.202$\pm$0.016} &  \textbf{0.018$\pm$0.001}   \\

\hline

 \textbf{CIFAR10} &     \multicolumn{4}{c}{} \\
 ~-BASE &       0.834$\pm$0.005 & 0.023$\pm$0.000 & 0.523$\pm$0.016 & 0.049$\pm$0.003        \\
 
 ~-MCD &        0.867$\pm$0.002 & 0.019$\pm$0.000 & 0.396$\pm$0.003 &  0.017$\pm$0.005   \\

 ~-DEEP &      0.877$\pm$0.003 & 0.017$\pm$0.000 & \textbf{0.365$\pm$0.015} &  \textbf{0.015$\pm$0.003}      \\
    
 ~-EE-ensemble &        0.854$\pm$0.001 & 0.021$\pm$0.000 & 0.446$\pm$0.011 & 0.033$\pm$0.001        \\

  ~-HYDRA &        0.818$\pm$0.004 & 0.026$\pm$0.000 & 0.632$\pm$0.017 & 0.069$\pm$0.001         \\
  
 ~-\qute &        0.858$\pm$0.001 & 0.020$\pm$0.000 & 0.428$\pm$0.019 & 0.025$\pm$0.003        \\

  ~-\qute$+$ &        \textbf{0.878$\pm$0.003} &	\textbf{0.017$\pm$0.000}	& \underline{0.369$\pm$0.008}	&0.026$\pm$0.001
 \\

\hline 

 \textbf{TinyImagenet}  & \multicolumn{4}{c}{} \\
 ~-BASE        & 0.351$\pm$0.005 & 0.004$\pm$0.000 & 5.337$\pm$0.084 &  0.416$\pm$0.003    \\
 
 ~-MCD      & 0.332$\pm$0.004 & 0.003$\pm$0.000 & 2.844$\pm$0.028 &  0.061$\pm$0.005    \\
 
 ~-DEEP    & 0.414$\pm$0.006 & 0.003$\pm$0.000 & 3.440$\pm$0.049 &   0.115$\pm$0.003   \\
 
 ~-EE-ensemble       & \textbf{0.430$\pm$0.005} & 0.003$\pm$0.000 & \textbf{2.534$\pm$0.046} &  \textbf{0.032$\pm$0.006}    \\

 ~-HYDRA       &   0.376$\pm$0.004 & 0.004$\pm$0.000 & 3.964$\pm$0.036 &  0.328$\pm$0.004    \\
 
 ~-\qute           &  0.395$\pm$0.014 & 0.004$\pm$0.000 & 3.700$\pm$0.123 &  0.282$\pm$0.009    \\    

 ~-\qute$+$ &      0.381$\pm$0.010	& \textbf{0.003$\pm$0.000}	& \underline{2.757$\pm$0.044} &	0.122$\pm$0.008

\end{tabular}
}
}
\caption{Calibration Metrics for all baselines on ID data. For F1, higher is better, and for BS and NLL, lower is better. The best results are marked in bold. All results are mean $\pm$ std-dev for 3 independent splits of ID test data. \vspace{-0.5cm}}
\label{table:uq}
\end{table}

%% file: figs/table-ev-effectiveness.tex
\begin{table*}[t]
\centering
\small
\scalebox{0.8}{

{

\begin{tabular}{l | c | c || c | c || c | c}
\multirow{2}{*}{\textbf{Model}} &  \multicolumn{2}{c||}{\textbf{F1 ($\uparrow)$}} & \multicolumn{2}{c||}{\textbf{BS ($\downarrow)$}} & \multicolumn{2}{c}{\textbf{NLL ($\downarrow)$}}\\
& \textbf{EV} & \textbf{no-EV}    & \textbf{EV} & \textbf{no-EV}   & \textbf{EV} & \textbf{no-EV}\\
\hline
CIFAR10 &  0.858$\pm$0.001 & 0.858$\pm$0.000     &   \textbf{0.0205$\pm$0.000} & 0.0206$\pm$0.000    &    \textbf{0.435$\pm$0.007}  & 0.459$\pm$0.006\\
TinyImagenet &  \textbf{0.380$\pm$0.011} & 0.35$\pm$0.004     &   \textbf{0.0043$\pm$7.2E-5} & 0.0046$\pm$5.7E-5    &    \textbf{3.813$\pm$0.105}  & 4.743$\pm$0.111\\
\hline

\end{tabular}
}
    }
\caption{Effectiveness of \emph{early-view}  assistance. Results averaged over 3 independent training runs.\vspace{-5mm}}
\label{table:ev-effectiveness}
\end{table*}

%% file: figs/table-TS.tex
\begin{table}[t]
\centering
\small
\scalebox{0.7}{

{
\begin{tabular}{l | c c c  }
\textbf{Model}  & \textbf{F1 ($\uparrow$)}  & \textbf{BS ($\downarrow$)} & \textbf{NLL ($\downarrow$)} \\
\hline
\hline

\textbf{MNIST}  & \multicolumn{3}{c}{} \\
  ~-~BASE-TS &     0.937$\pm$0.002   &  	0.009$\pm$0.000 	  &   0.203$\pm$0.005	          \\
 
~-~QUTE-TS &  \textbf{0.942$\pm$0.001}	 &  \textbf{0.008$\pm$0.000}	 & \textbf{0.191$\pm$0.005}	  	\\

 ~-~QUTE &  0.941$\pm$0.004 & 0.009$\pm$0.000 & 0.199$\pm$0.010        \\

\hline

\textbf{SpeechCmd} & \multicolumn{3}{c}{}     \\   

  ~-~BASE-TS &  0.923$\pm$0.007	  & 0.009$\pm$0.000	  & 0.229$\pm$0.017	  	  			\\

~-~QUTE-TS &  0.926$\pm$0.006	&   0.009$\pm$0.000	&   0.233$\pm$0.010					\\

 ~-~QUTE &      \textbf{ 0.933$\pm$0.006} & \textbf{0.008$\pm$0.000} & \textbf{0.202$\pm$0.016} \\

 \hline

\textbf{CIFAR10} & \multicolumn{3}{c}{}    \\   

  ~-~BASE-TS &  0.834$\pm$0.000 & 	 0.023$\pm$0.000 	&  0.493$\pm$0.000   \\

~-~QUTE-TS &  0.852$\pm$0.002	 & 0.021$\pm$0.000	 & 0.444$\pm$0.013
					\\

 ~-~QUTE &      \textbf{0.858$\pm$0.001} & \textbf{0.020$\pm$0.000} & \textbf{0.428$\pm$0.019}    \\

\end{tabular}
}
}
\caption{Calibration with temperature scaling (TS) on ID}
\label{table:uq-ts}
\end{table}

%% file: figs/table-cid-detect-ts.tex
\begin{table}[t]
\centering
\small
\scalebox{0.7}{

{

\begin{tabular}{l | c | c | c}
 \textbf{AUPRC} ($\uparrow$)&   \textbf{MNIST-C} & \textbf{SpeechCmd-C} & \textbf{CIFAR10-C} (Severity 1 to 5)\\
\hline

\hline
~-~BASE-TS & 0.47  &	0.52 & 0.30, 0.42, 0.38, 0.50, 0.61
 \\ 	

~-~QUTE-TS & 0.52  &	0.53  & 0.26, 0.45, 0.51, 0.65, 0.76
\\ 	
~-~QUTE & \textbf{0.63}  &	\textbf{0.61} &   \textbf{0.30, 0.47, 0.52, 0.65, 0.78} \\ 	

\hline

\end{tabular}
}
}
\caption{Accuracy-drop detection with temperature-scaling}
\label{table:cid-ts}
\end{table}

%% file: figs/table-uq-postnet.tex
\begin{table}[t]
\centering
\small
\scalebox{0.7}{

{

\begin{tabular}{l | c | c | c }
\textbf{Model} &  \textbf{F1 ($\uparrow$)}  & \textbf{BS ($\downarrow$)} & \textbf{NLL ($\downarrow$)}  \\
\hline
\hline

MNIST-PostN &           0.92 & 0.012 & 0.286\\ 
MNIST-\qute &           \textbf{0.94} & \textbf{0.009} & \textbf{0.199}\\ 
\hline
CIFAR-PostN &           0.84 & 0.022 & 0.462\\ 
CIFAR-\qute &           \textbf{0.858} & \textbf{0.020} & \textbf{0.428}\\ 

\end{tabular}
}
}
\caption{Calibration comparison with Postnets on ID}
\label{table:uq-postn}
\end{table}

%% file: tex/7-conclusion.tex
\section{Conclusion and Discussion}
\label{sec:conclusion}
Uncertainty quantification is essential for model monitoring in tinyML, yet many prior works fail to address the resource constraints inherent in this domain.
We propose a novel resource-efficient ensemble architecture for uncertainty estimation that is optimized for tinyML.
At the core of our methodology is our finding that model overconfidence decreases with its size. We leverage the better uncertainty estimation quality of early-exits (EE) by injecting EE weights into the multiple lightweight classification heads created at the output.
\qute provides reliable uncertainty estimates in a single forward pass with \textbf{59\% smaller models} on average and a \textbf{31\% average reduction in latency} on low-power MCUs.
Furthermore, \qute effectively distills EE knowledge into the final exits, enabling it to capture both aleatoric and epistemic uncertainties more effectively. This capability leads to superior performance in detecting accuracy-drop events (corruptions) across a diverse range of datasets and model complexities. 
\qute serves as an excellent and cost-effective \emph{accuracy-monitoring} mechanism for field-deployed models.

\vspace{-1mm}
\textbf{\textit{Limitations/Discussions}}~The ensemble size ($K$) is restricted by the depth of the base network, which may not provide uncertainty estimates of sufficient quality for a given safety-critical application. In addition, we find that increasing $K$ beyond a certain point negatively influences the shared base network, leading to degradation in calibration (ablation study in Appendix~\ref{subsec:apdx-ee-v-uncertainty}). Furthermore, while \qute achieves the best calibration on tiny-sized models, it requires additional learning layers on large models and datasets. Alternatively, methods that improve early-exit calibration~\cite{jazbec2024-anytime-ee} could potentially improve effectiveness of \qute's EV-assistance. Interestingly, we observe that excessively increasing EV-assistance, such as by raising the depth multiplier in \qute’s output blocks, negatively impacts accuracy despite improving calibration.
However, its worth noting that the training times with \qute are higher compared to other methods (except HYDRA) due to weight transfer between train-batches. 
Finally, despite \qute's versatility/effectiveness in OOD detection, specialized OOD detectors should be considered, especially for large models.


\vspace{-1mm}



%% file: tex/9-appendix.tex
\section{Appendix}
\label{sec:apdx}

\subsection{Training and dataset details}
\label{subsec:apdx-train}
In this section, we describe the details of the models evaluated and specifics of training and the experimental setup.

\subsubsection{Datasets}
In our evaluations, we use four in-distribution datasets for training all baselines methods we evaluate: 1)~MNIST~\cite{mnist}, 2)~SpeechCommands~\cite{speech_commands}, 3)~CIFAR10~\cite{cifar-10} and 4)~TinyImagent~\cite{le2015-tinyimagenet}. 

\textbf{MNIST}~
MNIST is a dataset of handwritten digits containing 60,000 grayscale images of size 28$\times$28 and a test set of 10,000 images. MNIST contains 10 classes

\textbf{SpeechCommands}~
Speech-Cmd is a collection of short audio clips, each spanning 1 second. The dataset consists of utterances for 35 words and is commonly used for benchmarking keyword spotting systems. We train our model (DSCNN) to recognize ten words out of 35: \textit{Down, Go, Left, No, Off, On, Right, Stop, Up, Yes}. Thus, the number of classes is 10. The audio files in WAV format are preprocessed to compute Mel-frequency cepstral coefficients (mel-spectograms). The mel-spectograms are of size 49$\times$10 with a single channel. The DSCNN model is trained for 10 epochs with a batch size of 100.

\textbf{CIFAR10}~
CIFAR10 dataset consists of 60,000 32$\times$32 rgb images out of which 10,000 images are in the test set. It contains 10 classes and thus 6000 images per class.

\textbf{TinyImagenet}~
TinyImagenet is a smaller version of the Imagenet dataset containing 200 classes instead of 1000 classes of the original Imagenet. Each class in TinyImagenet has 500 images in the train set and the validation set contains 50 images per class. The size of the images are resized and fixed at 64$\times$64$\times$3.

For image classification tasks, we use the following corrupted versions of ID: 1)~MNIST-C~\cite{mu2019-mnistc}, 2)~CIFAR10-C and 3)~TinyImagenet-C~\cite{hendrycks2019-benchmarking-corruptions}. All corruptions are drawn from  4 major sources: noise, blur, weather and digital. The \textit{digital} and \textit{blur} corruptions induce high epistemic uncertainty, whereas, \textit{noise} corruptions induce higher aleatoric uncertainty. In contrast, \textit{weather} corruptions tend to encompass both epistemic and aleatoric components. Most corruptions are systematic transformations of input images, which predominantly introduce epistemic uncertainty. Pure noise corruptions are the primary source of aleatoric uncertainty; however, many corruptions exhibit both components, often leaning toward epistemic.

\textbf{MNIST-C}~
The MNIST-C dataset contains 15 corrupted versions of MNIST - \textit{shot\_noise, impulse\_noise, glass\_blur, fog, spatter, dotted\_line, zigzag, canny\_edges, motion\_blur, shear, scale, rotate, brightness, translate, stripe, identity}. All the corruptions are of a fixed severity level. 

\textbf{CIFAR10-C}~
CIFAR10-C includes 19 different types of corruptions with 5 severity levels which gives us 19$\times$5$=$95 corrupted versions of CIFAR-10 ID. We use all 95 corrupted versions in our experiments. The list of corruptions are: \textit{gaussian\_noise, brightness, contrast, defocus\_blur, elastic, fog, frost, frosted\_glass\_blur, gaussian\_blur,  impulse\_noise,  jpeg\_compression, motion\_blur, pixelate, saturate, shot\_noise,  snow, spatter, speckle\_noise, zoom\_blur}.

\textbf{TinyImagenet-C}~
TinyImagenet-C includes 15 different types of corruptions with 5 severity levels which gives us 15$\times$5$=$75 corrupted versions of ID. We use all 75 corrupted versions in our experiments. The list of corruptions are: \textit{gaussian\_noise, brightness, contrast, defocus\_blur, elastic\_transform, fog, frost, glass\_blur, impulse\_noise,  jpeg\_compression, motion\_blur, pixelate, shot\_noise,  snow, zoom\_blur}.

\textbf{SpeechCmd-C}~For the audio classification task of Keyword spotting on SpeechCmd, we use corrupt the audio using the audiomentations library~\cite{jordal2024-audiomentations} with the following 11 corruptions: \textit{gaussian noise, air absorption, band pass filter, band stop filter, high pass filter, high shelf filter, low pass filter, low shelf filter, peaking filter, tanh distortion, time mask, time stretch}.

\subsubsection{Training details}
We evaluate four models in our experiments:1)~4-layer CNN on MNIST, 2)~4 -layer Depthwise-separable CNN (DSCNN) on SpeechCmd, 3)~Resnet-8 with 3 residual stacks from the MLPerf Tiny benchmark suite~\cite{mlperf-tiny-benchmark} on CIFAR10 and 4)~MobilenetV2~\cite{mobilenets} on TinyImagenet.

The 4-layer CNN is trained for 20 epochs with a batch size of 256, the DSCNN model is trained for 10 epochs with a batch size of 100, the Resnet-8 model is trained for 200 epochs with batch size of 32 and the MobilenetV2 model is trained for 200 epochs with a batch size of 128. All models are trained with Adam optimizer with momentum of 0.9 and an initial learning rate of 0.001, expect DSCNN on SpeechCmd which uses an initial learning rate of 0.0005. The learning rate is decayed by a factor of 0.99 every epoch for all image classification tasks. We follow a step function for SpeechCmd that reduces learning rate by half every 2 epochs. For models with \qute architecture, we achieve the weight transfer from early-exits to all $\{f_{out}(.)\}_{k=1}^{K}$ using a callback that sets the weights of each $f_{out_k}$ with weights copied from the corresponding assisting early-exit at the beginning of each training batch.

\subsection{Weighting the loss at EV-exits}~
As described in Section~\ref{sec:design}, we weight the loss at EV-exits with a higher factor $w_{EV_k}$ such that $w_{EV_k} = w_{EV_{k-1}} + \delta$. This is done to further promote diversity across the EV-exits and to vary the influence of each EV-exit on the shared network's weights. Utilizing distinct weighting factors across EV-exits implies that each EV-exit contributes to the final loss calculation with varying degrees of influence. Consequently, this diversity extends to the weight updates of the shared base network, thereby injecting diversity into the classification heads. We empirically set $\delta$ to be 0.5. To determine $w_{EV_0}$, we repeat training with $w_{EV_0}$ set to $\{2,3,4,5\}$. We find that for $w_{EV_0}>3$, the NLL starts dropping steadily because the higher loss at EV-exits overshadow the losses of the early-exits. This diminishes the influence of early-exits at EV-exits, which is undesirable. Hence, we set $w_{EV_0} = 3$ in all our experiments.

\begin{figure}
    \centering
    \scalebox{0.7}{
    \includegraphics[width=0.9\linewidth]{ 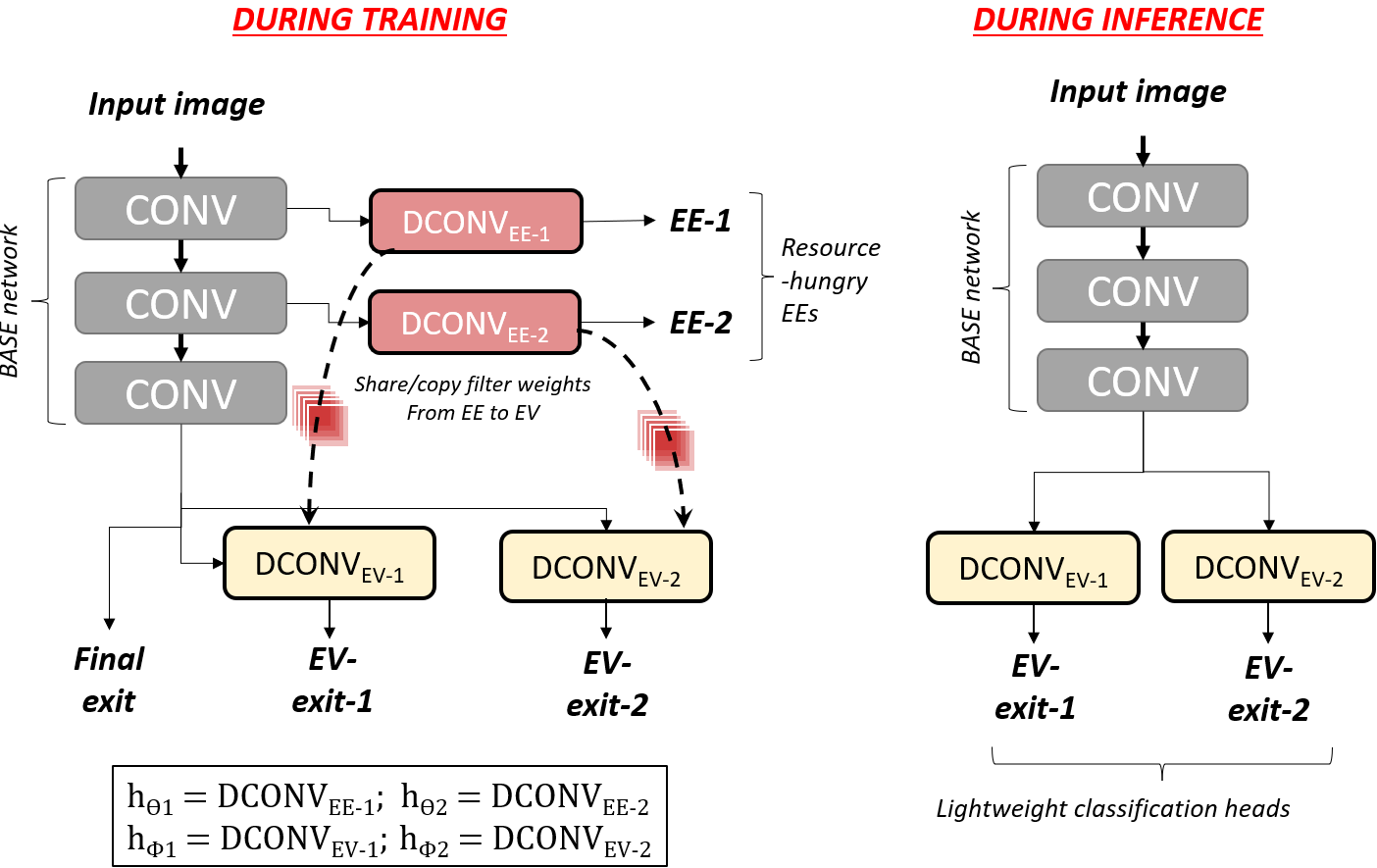}
    }
    \caption{Depiction of QUTE during training and inference. The early-exits (EE) are only used during training for weight-transfer. During inference, the EEs are removed, and only EV-exits are retained. We do not include other layers such as dense and pointwise convolution in the figure for brevity.}
    \label{fig:qute-visual}
\end{figure}

\subsection{Corrupted-in-distribution datasets}
We construct the corrupted-in-distribution (CID) datasets for our evaluations as follows. For MNIST-C with a fixed severity level, we append each corrupted version of MNIST to the MNIST-ID dataset to create 15 ID+CID datasets. For CIFAR10-C and TinyImagenet-C, which have 19 and 15 different corruptions respectively with 5 severity levels each, we construct the corrupted datasets as follows. For each type of corruption, we randomly select $p$ samples from each of the 5 severity levels. Next, we concatenate all these samples to create a new corrupted dataset of size $5\times p$. $p$ is selected such that $5\times p$ = size of ID dataset. This process is repeated for all corruptions. The datasets thus obtained contain samples from all severity levels. For example, for CIFAR10-C which consists of 19 corruptions, this process yields 19 ID+CID datasets.




\subsection{Accuracy-drop detection experiments}
\label{subsec:apdx-cid-results}

For experiments to detect accuracy drop/CID detection described in Section~\ref{subsec:results-cid}, we append the ID dataset with each corrupted dataset obtained from the methodology described above. For example, for experiments with CIFAR10-C, we obtain 19 ID+CID datasets that contains both  ID samples and corrupted samples (from all severity levels). 
Next, as described in Section~\ref{sec:methodology}, for each prior work we evaluate, we first iterate over \emph{only} ID and obtain predictions from the model while computing the moving average of accuracy of the past $m$ predictions using a sliding-window. In this way, we obtain the accuracy distribution of the sliding-window $\mathcal{A}_{SW}$ on ID, and then compute its mean $\mu_{ID}$ and  standard-deviation $\sigma_{ID}$. Next, we iterate over all ID+CID datasets while computing the moving average of confidence of the past $m$ predictions using a sliding-window ($\mathcal{C}_{SW}$). We record all instances of $\mathcal{C}_{SW}$ dropping below a certain threshold $\rho$. These events are denoted as CID-predicted events. At the same time, we also compute $\mathcal{A}_{SW}$ for evaluating how many CID-predicted events are  actually a CID-detected event (true positive).
Finally, the description of true positive (TP), false positives (FP) , false negatives (FN) and true negatives (TN) are as follows.
\begin{itemize}
    \item True positive: ~$\mathcal{C}_{SW} < \rho$ and ~$\mathcal{A}_{SW} \leq \mu_{ID} - 3\cdot\sigma_{ID}$ 
    \item False positive: ~$\mathcal{C}_{SW} < \rho$ and ~$\mathcal{A}_{SW} > \mu_{ID} - 3\cdot\sigma_{ID}$
    \item True negative: ~$\mathcal{C}_{SW} > \rho$ and ~$\mathcal{A}_{SW} > \mu_{ID} - 3\cdot\sigma_{ID}$ 
    \item False negative: ~$\mathcal{C}_{SW} > \rho$ and ~$\mathcal{A}_{SW} \leq \mu_{ID} - 3\cdot\sigma_{ID}$
\end{itemize}

In this way, we collect the TP, FP, TN, FN from each ID+CID datasets and compute the average precision and recall for a given threshold $\rho$. Finally, we report the AUPRC. We choose AUPRC because the precision-recall curve is resistant to imbalance in datasets.

\section{Ablation Studies}
\label{sec:apdx-ablation}

\subsection{Uncertainty quantification vs Ensemble size}
\label{subsec:apdx-ee-v-uncertainty}
In this section, we conduct an ablation study to investigate the effect of number of early-exits on uncertainty estimation quality.
The ensemble size $|\mathcal{K}|$ is an hyperparameter that depends on the computation/resource budget, and it is bounded above by the depth of the base network. We vary the ensemble size and investigate its effect on calibration quality in MobilenetV2.
\begin{figure}[h]
    \centering
    \hspace{-2cm}
    \begin{subfigure}[]{\textwidth/2}
        \centering
        \scalebox{0.6}{\input{figs/ee-v-uq-acc}}
        \label{fig:ee-v-uq-acc}
        \caption{Accuracy ($\uparrow$)}
    \end{subfigure}
    \begin{subfigure}[]{\textwidth/2}
        \centering
        \scalebox{0.6}{\input{figs/ee-v-uq-nll}}
        \label{fig:ee-v-uq-nl}
        \caption{NLL ($\downarrow$)}
    \end{subfigure}
    \caption{Effect of Ensemble size on Accuracy and NLL in MobilenetV2 on TinyImagenet. Red dotted horizontal line indicates accuracy and NLL of BASE in respective plots}
    \label{fig:ee-v-acc_nll}
\end{figure}
Figure~\ref{fig:ee-v-acc_nll} shows the effect on accuracy and NLL in MobilnetV2 for $|\mathcal{K}|$ ranging from 2-10. The red line shows the accuracy and NLL of base network in respective plots. As shown, the accuracy does not always improve with ensemble size. This demonstrates that there is a limit to the improvement in accuracy due to early-exits. On the other hand, NLL steadily improves as ensemble size increases with the least NLL obtained for $|\mathcal{K}|=8$. For $|\mathcal{K}|=10$, the NLL rises slightly. Our investigations revealed that since the base network, early-exits and the early-view (EV) exits are part of the same network architecture, and are trained simultaneously, the disruption in base network's weights due to the losses of early-exits and EV-exits reaches a tipping-point after a certain ensemble size thereby, causing a drop off in performance.

\subsection{Ablations with EE-ensemble}
\label{subsec:apdx-ee-ensemble}

\subsubsection{Excluding final exit from final prediction}

As described in Section~\ref{sec:methodology}, we exclude the original output block of the base network when computing the final prediction vector for \qute because we find it that is overconfident and harms calibration quality. However, we include the original output block's prediction vector along with that of of all early-exits for computing the final prediction vector for EE-ensemble, consistent with the original paper~\citet{qendro2021-EEensemble}. We conduct an ablation study on CIFAR10 using Resnet-8, where like \qute, we exclude the original output block's prediction vector when computing the final prediction vector. Table~\ref{table:uq-ee-ensemble-nofout} reports the calibration metrics on CIFAR10-ID for two configurations: 1)~including original output block in computing final prediction vector and 2)~excluding original output block in computing final prediction vector. 
\input{figs/table-ee-ensemble-nofout}
As seen, the removal of original output block negatively impacts EE-ensemble leading to poor accuracy and calibration. Therefore, we include the original output block in computing the final prediction vector in our main results for EE-ensemble.

\subsubsection{Using \qute's output block architecture for early-exits in EE-ensemble}

\citet{qendro2021-EEensemble} adds \emph{additional fully-connected layers} at the early-exit for EE-ensemble to match the learning capacities of all exits. However, this leads to a large overhead (Table~\ref{table:uq}). In this ablation study, we show that the additional layers at the early-exit are necessary for EE-ensemble to achieve good accuracy and calibration. We replace the resource-hungry early-exits of EE-ensemble with the architecture of \qute's output blocks. 
\input{figs/table-ee-ensemble-barebones}
Table~\ref{table:uq-ee-ensemble-barebones} reports the calibration metrics for two configurations: 1)~EE-ensemble \emph{with additional FC layers} at the early-exit and 2)~EE-ensemble with \qute's lightweight output block architecture at the early-exits. As seen, the configuration with extra learning layers clearly outperforms the one without on both ID and CID in terms of accuracy and calibration. The early-exits placed at different depths work with much less information compared to the final exit. As a result, the accuracy and calibration invariably degrade if the learning capacities of all early-exits do not match that of the final exit.



\subsection{MCU results}
\label{subsec:apdx-mcu}
\input{figs/table-on-device-results}
Tables~\ref{table:mcu-big} and~\ref{table:mcu-small} reports the on-device results for \bigmcu and \smallmcu respectively. We report the accuracy, code-size and the latency of each method. On \bigmcu, all methods fit on the device. The low memory-footprint of \qute allows it reduce the code-size consistently across datasets, leading to lower latency and, consequently, reduced energy per prediction. In contrast, EE-ensemble has higher memory requirements due to the expensive dense layers at its early exits. Furthermore, with no parallel execution capabilities on an MCU, the latency of DEEP with two ensemble members takes roughly twice as long to execute compared to BASE. This issue could worsen significantly with a larger ensemble size.

We observe a similar trend on the \smallmcu, with \qute providing the best uncertainty quantification (UQ) per unit of resource used. However, both EE-ensemble and DEEP exceed the available 256KB of ROM on \smallmcu, making them unsuitable for ultra-low-power devices. In such scenarios, \qute emerges as the only practical solution for effective UQ. Furthermore, we observe an interesting trend on the MNIST dataset. When using a 4-convolution layer model (4-layer CONV), the code sizes of all methods are quite similar, even exceeding that of the SpeechCmd BASE model, which employs a 4-layer depthwise-separable convolutional model (4-layer DSCNN) with 20,000 more parameters.
A deeper analysis of the MAP files, which detail memory allocation, reveals that MNIST uses standard convolutions instead of separable convolutions. Therefore, the memory required for code is significantly higher for the 4-layer CONV model compared to the 4-layer DSCNN. In fact, the memory required for operators and code in the 4-layer CONV model is four times greater than the memory needed to store its weights.

These results highlight the efficiency of \qute. While other methods may struggle with memory limitations, \qute manages to deliver effective UQ without compromising on resource utilization. Its design allows for optimal performance even in constrained environments, making it a compelling choice for applications where both accuracy and efficiency are of paramount importance.

\section{Evaluation metrics}
\label{sec:apdx-metrics}
We evaluate uncertainty using the following metrics.

\textbf{Brier Score}~(lower is better):~It is a proper scoring rule that measures the accuracy of predictive probabilities. Incorrect predictions with high predictive confidence are penalized less by BS. Thus, BS is less sensitive to corruptions and incorrect predictions. Therefore, NLL is a better measure of uncertainty to compare with other methods.
\begin{eqnarray}
    BS = \frac{1}{N} \underset{l \in \{1,2,..L\}}{\sum_{n=1}^N}~~(p_\Theta(y=l|x) - \mathds{1}(y=l))^2  
\end{eqnarray}
where, $\mathds{1}$ is an identity function and $L$ is the number of classes.

\textbf{Negative log-likelihood}~(lower is better):~It is a proper scoring rule that measures how probable it is that the predictions obtained are from the in-distribution set. It depends on both the uncertainty (predictive confidence) and the accuracy of the predictions.
\begin{eqnarray}
    NLL = -  \underset{l \in \{1,2,..L\}}{\sum}~~\mathds{1}(y=l)\cdot logp_\Theta(y=l|x)  +  (1-\mathds{1}(y=l))\cdot~log(1-p(y=l|x))
\end{eqnarray}

\textbf{Expected Calibration Error (ECE)}~(lower is better):~ECE measures the absolute difference between predicted confidence and actual accuracy across different confidence intervals. The confidence is in the range [0,1]. ECE divides the confidence range into $M$ intervals/bins of size $\frac{1}{M}$ and computes the bin accuracies and confidence of each bin before averaging them to provide the final ECE score.
\begin{eqnarray}
    ECE = \sum_{m=1}^{M} \frac{|B_m|}{N}\cdot |acc(B_m) - conf(B_m)| \label{eq:ece}
\end{eqnarray}
where, $N$ is total number of predictions, $B_m$ is the $m$\textsuperscript{th} bin spanning the interval $(\frac{m-1}{M}, \frac{m}{M})$. A low ECE score is desirable, and indicates less disparity between confidence and accuracy across all intervals. However, the ECE score has several limitations and susceptibilities.

\textbf{Limitations of ECE}

\vspace{-1mm}
1)~ECE divides predicted probabilities into discrete bins. However, there might be a severe imbalance of samples across the bins because neural networks tend to always predict with high confidence~\cite{nixon2019-measuring-calib}. This causes only a few bins (concentrated towards the high confidence region) to contribute the most to ECE.
In addition, ECE is also sensitive to the bin boundaries i.e., the number of bins.

\vspace{-1mm}
2)~ECE does not differentiate between specific types of miscalibration such as overconfidence/underconfidence , often providing a simplistic and (overly) optimistic view of the calibration quality. Therefore, ECE is not a proper scoring metric. A model might have a lower ECE without genuinely having good calibration.

For these reasons, we do not emphasize ECE in our main results as they are not indicative of the model's true calibration quality. We focus more on proper scoring metrics like NLL and BS in our main results.

%% file: figs/ee-v-uq-acc.tex
\begin{tikzpicture}
	\pgfplotstableread{
	ensemble_size   accuracy 	
	    2	0.3609
            4	0.3717
            6	0.3605
            8	0.371
            10	0.356
	}\data
		\begin{axis}[
			width=\textwidth+1.5cm,
			height=\graphHeight*2-1cm,
			%
			major tick length=3pt,
			%
			major grid style={dashed,color=gray!50},
			minor grid style={color=gray!50},
			ymajorgrids=true,
			xmajorgrids=true,
			%
			xticklabel style={text height=5pt,font=\scriptsize},
			%
			xlabel = Ensemble Size ($|\mathcal{K}|$),
			xlabel style={font=\scriptsize},
		    xlabel near ticks,
                xmin = 2,
                xmax = 10,
		   extra x ticks={0,2,...,10},
			%
			ymin=0.34,ymax=0.38,
			yticklabel style={font=\scriptsize},
			%
			ylabel style={align=center,font=\scriptsize},
			ylabel=Accuracy (\%),
               legend style={at={(2,2)},anchor=north east,draw=none, cells={align=left}, font=\scriptsize,text width=0.5cm},
			legend columns=1,
                legend pos= north east,
			%
			enlarge x limits=0.1,
			%
		]

		\addplot+[bblue, smooth,] table[x=ensemble_size, y=accuracy] {\data};
        \draw [loosely dashed, thick, red] (0,0.351) -- (12,0.351);
	
		
		


        
		\end{axis}
	\end{tikzpicture}

%% file: figs/ee-v-uq-nll.tex
\begin{tikzpicture}
	\pgfplotstableread{
	ensemble_size   nll 	
	    2	     4.43
            4	      4.24
            6	      4.21
            8	      3.82
            10	     4.03            
	}\data
		\begin{axis}[
			width=\textwidth+1.5cm,
			height=\graphHeight*2-1cm,
			%
			major tick length=3pt,
			%
			major grid style={dashed,color=gray!50},
			minor grid style={color=gray!50},
			ymajorgrids=true,
			xmajorgrids=true,
			%
			xticklabel style={text height=5pt,font=\scriptsize},
			%
			xlabel = Ensemble Size ($|\mathcal{K}|$),
			xlabel style={font=\scriptsize},
		    xlabel near ticks,
                xmin = 2,
                xmax = 10,
		   extra x ticks={2, 4,...,10},
			%
			ymin=3.5,ymax=5.7,
			yticklabel style={font=\scriptsize},
			%
			ylabel style={align=center,font=\scriptsize},
			ylabel=NLL,
               legend style={at={(2,2)},anchor=north east,draw=none, cells={align=left}, font=\scriptsize,text width=0.5cm},
			legend columns=1,
                legend pos= north east,
			%
			enlarge x limits=0.1,
			%
		]

		\addplot+[ppurple, smooth] table[x=ensemble_size, y=nll] {\data};
	 \draw [loosely dashed, thick, red] (0,5.337) -- (12,5.337);
		
		


        
		\end{axis}
	\end{tikzpicture}

%% file: figs/table-ee-ensemble-nofout.tex
\begin{table*}[h]
\centering
\small
\scalebox{1}{
{

\begin{tabular}{l | c c c || c c c }
 & \multicolumn{3}{|c||} {\textit{Including final-exit}} & \multicolumn{3}{c}{\textit{Excluding final-exit}}\\
\hline
\textbf{Resnet-8} &  \textbf{F1 ($\uparrow$)}  & \textbf{BS ($\downarrow$)} & \textbf{NLL ($\downarrow$)}  &    \textbf{F1 ($\uparrow$)}  & \textbf{BS ($\downarrow$)} & \textbf{NLL ($\downarrow$)}\\
\hline

 \textbf{CIFAR10} & \multicolumn{3}{c||}{} & \multicolumn{3}{c}{}\\
 ~-~EE-ensemble &       0.854 & 0.021 & 0.446 &       0.818 & 0.0026 & 0.561    \\
 
\end{tabular}
}
}
\caption{Calibration Metrics for Resnet-8 on CIFAR10-ID computed \emph{including} original output block and \emph{excluding} original output block from computation of final prediction vector.}
\label{table:uq-ee-ensemble-nofout}
\end{table*}

%% file: figs/table-ee-ensemble-barebones.tex
\begin{table*}[h]
\centering
\small
\scalebox{1}{
{

\begin{tabular}{l | c c c || c c c }
 & \multicolumn{3}{|c||} {\textit{In-distribution}} & \multicolumn{3}{c}{\textit{Corrupted-in-distribution}}\\
\hline
\textbf{Resnet-8} &  \textbf{F1 ($\uparrow$)}  & \textbf{BS ($\downarrow$)} & \textbf{NLL ($\downarrow$)}  &    \textbf{F1 ($\uparrow$)}  & \textbf{BS ($\downarrow$)} & \textbf{NLL ($\downarrow$)}\\
\hline

 \textbf{CIFAR10} & \multicolumn{3}{c||}{} & \multicolumn{3}{c}{}\\
 ~-~EE-ensemble \textit{w/ additional layers}&       0.85 & 0.021 & 0.446 &       0.64 & 0.0049 & 1.256    \\
 ~-~EE-ensemble \textit{w/o additional layers}&       0.78 & 0.031 & 0.661 &       0.57 & 0.0058 & 1.459    \\
 
\end{tabular}
}
}
\caption{Calibration Metrics for Resnet-8 on CIFAR10-ID computed for EE-ensemble \emph{with additional FC layers} and \emph{without additional FC layers}}
\label{table:uq-ee-ensemble-barebones}
\end{table*}

%% file: figs/table-on-device-results.tex
\begin{table}[h!]
\centering
\small
\scalebox{0.8}{
{

\begin{tabular}{l | c c c || c c c }
 &  \multicolumn{3}{|c||} {\textit{SpeechCmd}} & \multicolumn{3}{c}{\textit{CIFAR10}}\\
\hline
\textbf{Model} &  \textbf{Accuracy} & \textbf{Size (KB)}  & \textbf{Latency (ms)} & \textbf{Accuracy} & \textbf{Size (KB)}  & \textbf{Latency (ms)}  \\
\hline

 ~-~BASE &  0.92 & 294.5 & 22.7 &       0.83 & 344.4 & 58.04     \\

 ~-~DEEP &  0.93 & 321.6 & 45.4 &       0.88 & 423.4 & 116.2       \\

 ~-~EE-ensemble &  0.92 & 361.5 & 51.7 &       0.85 & 735.5 & 65.4       \\

 ~-~\qute &  0.93 & 297.8 & 24.07 &     0.85 & 355.3 & 59.6     \\
\end{tabular}
}
}
\caption{Microcontroller results on \bigmcu (STM32F767ZI)}
\label{table:mcu-big}
\end{table}

\begin{table}[h!]
\centering
\small
\scalebox{0.8}{
{

\begin{tabular}{l | c c c || c c c || c c c}
 &  \multicolumn{3}{|c||} {\textit{SpeechCmd}} & \multicolumn{3}{c||}{\textit{CIFAR10}}  & \multicolumn{3}{c}{\textit{MNIST}}\\
\hline
\textbf{Model} &  \textbf{Accuracy} & \textbf{Size (KB)}  & \textbf{Latency (ms)} & \textbf{Accuracy} & \textbf{Size (KB)}  & \textbf{Latency (ms)}  &  \textbf{Accuracy} & \textbf{Size (KB)}  & \textbf{Latency (ms)}\\
\hline

 ~-~BASE &  0.92 & 84.8 & 160.7 &       					0.83 & 187.5 & 291.2     		& 0.906 & 100.3 & 5.01\\

 ~-~DEEP &  0.93 & 111.3 & 321.2 &       				\multicolumn{3}{c||}{\texttt{DNF/OOM}}        					& 0.93 & 109.3 & 9.3  \\

 ~-~EE-ensemble &  0.92 & 157.7 & 376.3 &       		\multicolumn{3}{c||}{\texttt{DNF/OOM}}        				& 0.93 & 113.5 & 23.7  \\

 ~-~\qute &  0.93 & 93 & 173.1 &     				0.85 & 201 & 298.05       				& 0.92 & 108.2 & 5.8  \\
\end{tabular}
}
}
\caption{Microcontroller results on \smallmcu (STM32L432KC). \texttt{DNF/OOM} indicates \textit{did-not-fit/out-of-memory}}
\label{table:mcu-small}
\end{table}